\documentclass[runningheads]{llncs}
\usepackage[pagebackref=true,breaklinks=true,colorlinks,bookmarks=false]{hyperref}
\usepackage{graphicx}
\usepackage{amsmath,amssymb} 
\usepackage[utf8]{inputenc}

\usepackage{float}
\usepackage{bm}
\usepackage{multirow}

\usepackage{tabularx}
\usepackage{algorithm,algpseudocode}
\usepackage[symbol]{footmisc}
\usepackage{placeins}
\usepackage{booktabs}

\usepackage{pifont} 

\usepackage{adjustbox}

\usepackage[mode=buildnew]{standalone}
\usepackage{tikz}
\usetikzlibrary{positioning}
\usetikzlibrary{decorations.pathreplacing}
\usetikzlibrary{shapes,snakes}
\usetikzlibrary{calc}
\usetikzlibrary{backgrounds}

\begin{document}

\makeatletter 
\newcommand{\printfnsymbol}[1]{%
	\textsuperscript{\@fnsymbol{#1}}%
} 
\makeatother
\renewcommand{\thefootnote}{\fnsymbol{footnote}}

\newcommand{\cmark}{\ding{51}}%
\newcommand{\xmark}{\ding{55}}%

\pagestyle{headings}
\mainmatter

\def\ACCV20SubNumber{***}  
\title{FreezeNet: Full Performance by Reduced Storage Costs} 
\titlerunning{FreezeNet: Full Performance by Reduced Storage Costs}
%
\author{Paul Wimmer\inst{1,2}\thanks{equal contribution}\orcidID{0000-0002-0503-4539} 
	\and
Jens Mehnert\inst{1}\printfnsymbol{1}\orcidID{0000-0002-0079-0036} \and
Alexandru Condurache\inst{1,2}}
\authorrunning{P. Wimmer, J. Mehnert and A. Condurache}
%
\institute{Robert Bosch GmbH, Daimlerstrasse 6, 71229 Leonberg, Germany\\\email{\{Paul.Wimmer,JensEricMarkus.Mehnert,AlexandruPaul.Condurache\}@de.bosch.com}\and
University of Luebeck, Ratzeburger Allee 160, 23562 Luebeck, Germany}

\maketitle

\setcounter{footnote}{0} 
\renewcommand{\thefootnote}{\arabic{footnote}}
\begin{abstract}

	Pruning generates sparse networks by setting parameters to zero.
	In this work we improve one-shot pruning methods, applied before training,
	without adding any additional storage costs while preserving the sparse
	gradient computations. The main difference to pruning is that we do
	not sparsify the network's weights but learn just a few key parameters
	and keep the other ones fixed at their random initialized value. This
	mechanism is called \emph{freezing the parameters}. Those frozen
	weights can be stored efficiently with a single $32$bit random seed
	number. The parameters to be frozen are determined one-shot by a
	single for- and backward pass applied before training starts. We call
	the introduced method \emph{FreezeNet}. In our experiments we show
	that FreezeNets achieve good results, especially for extreme freezing
	rates. Freezing weights preserves the gradient flow throughout the
	network and consequently, FreezeNets train better and have an increased
	capacity compared to their pruned counterparts. On the classification tasks
	MNIST and CIFAR-$10$/$100$ 
	we outperform SNIP, in this setting the best reported
	one-shot pruning method, applied before training. On MNIST, FreezeNet achieves
	$99.2\%$ performance of the baseline LeNet-$5$-Caffe architecture, while compressing the number of trained and stored parameters
	by a factor of $\times157$.
	
	 \and {Backpropagation}
\end{abstract}

\section{Introduction}

Between $2012$ and $2018$, computations required for deep learning
research have been increased by estimated $300,000$ times which corresponds
to doubling the amount of computations every few months \cite{schwartz_2019}.
This rate outruns by far the predicted one by Moore's Law \cite{gustafson_2011}.
Thus, it is important to reduce computational costs and memory requirements
for deep learning while preserving or even improving the status quo
regarding performance \cite{schwartz_2019}.  

\emph{Model compression} lowers storage costs, speeds up inference
after training by reducing the number of computations, or accelerates
the training which uses less energy. A method combining these factors
is \emph{network pruning}. To follow the call for more sustainability
and efficiency in deep learning we improve the best reported pruning
method applied before training, SNIP (\cite{lee_2018} Single-shot
Network Pruning based on Connection Sensitivity), by freezing the
parameters instead of setting them to zero.

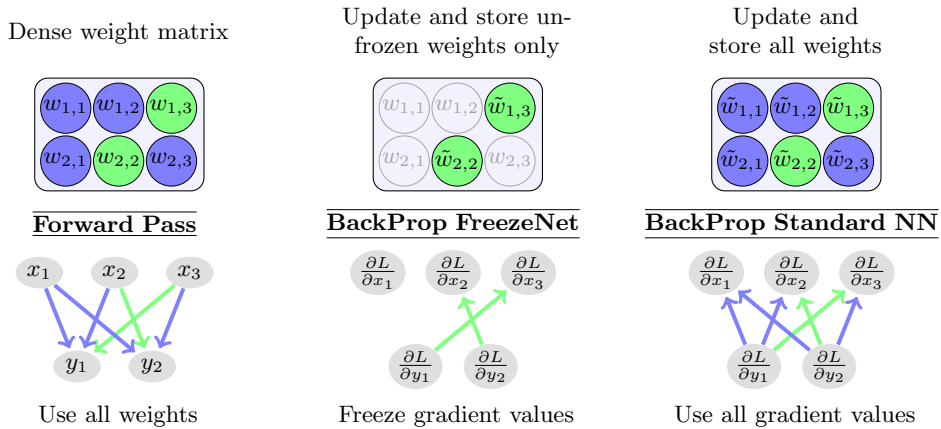
\begin{figure}[tb]
	\begin{center}
						\def\layersep{1.75cm}
				\begin{tikzpicture}[shorten >=1pt,->,draw=black!50, node distance=\layersep]     
				\tikzstyle{every pin edge}=[<-,shorten <=1pt];
				
				\tikzstyle{neuron}=[ellipse, minimum height=12pt,minimum width=18pt,fill=black!25,inner sep=0pt];     
				\tikzstyle{input neuron}=[neuron, fill=gray!25];     
				\tikzstyle{output neuron}=[neuron, fill=red!50];    
				\tikzstyle{hidden neuron}=[neuron, fill=gray!25];    
				
				\tikzstyle{annot} = [text width=12em, text centered];
				
				\tikzstyle{frozen} = [blue!50, line width=1.5pt];
				\tikzstyle{trained} = [green!50, line width=1.5pt];     
				\tikzstyle{trained} = [green!50, line width=1.5pt];
				\tikzstyle{trained back} = [green, line width=2pt]; 
				
				\tikzset{matrix/.style={black, rectangle,  draw=black, rounded corners=5, fill=blue!5, minimum width=2.25cm,minimum height=1.5cm, on grid}}
				
				\tikzstyle{weight}=[circle,fill=black!25,minimum size=19pt,inner sep=0pt,draw=black,line width=0.25pt];     
				\tikzstyle{standard weight}=[weight,fill=green!20];     
				\tikzstyle{frozen weight}=[weight,fill=blue!20];     
				\tikzstyle{frozen weight thinn}=[weight,fill=blue!5, text=black!35, draw=black!35];     
				\tikzstyle{frozen weight thick}=[weight,fill=blue!50];     
				\tikzstyle{standard weight thick}=[weight,fill=green!50];

				\def\vert{4.5};
				
				\foreach \name / \y in {1,...,3}         
				\node[input neuron] (I-\name) at (\y, 0) {$x_ \y$};
				
				\foreach \name / \y in {1,...,2}         
				\path[yshift=0.5cm]             
				node[hidden neuron] (H-\name) at (\y + .5, - \layersep) {$y_\y$};
				
				\begin{scope}
				\node [matrix] (m1) at (2.0625,1.8375) {};
				\node [frozen weight thick] (w11) at (1.35,  1.25 * \layersep) {$w_{1,1}$};
				\node [frozen weight thick] [right=0.75pt of w11] (w12){$w_{1,2}$};    
				\node [standard weight thick] [right=0.75pt of w12] (w13){$w_{1,3}$};    
				\node [frozen weight thick] [below=0.75pt of w11] (w21){$w_{2,1}$};    
				\node [standard weight thick] [right=0.75pt of w21] (w22){$w_{2,2}$};    
				\node [frozen weight thick] [right=0.75pt of w22] (w23){$w_{2,3}$};  
				\end{scope}
				
				\path (I-1) edge[frozen] (H-1);    
				\path (I-2) edge[trained] (H-2);    
				\path (I-3) edge[trained] (H-1);    
				\path (I-1) edge[frozen] (H-2);    
				\path (I-2) edge[frozen] (H-1);    
				\path (I-3) edge[frozen] (H-2);
				
				\node[annot,above of=I-2, node distance=.65cm] (hl1) {  $\overline{\textbf{\underline{Forward Pass}}}$};     
				\node[annot,above of=w12, node distance=1cm] (aa1) {\footnotesize Dense weight matrix};     
				\node[annot,below right=0.25cm and -1.75 cm of H-1, node distance=0.7cm] (ab1) {\footnotesize Use all weights}; 
				
				\foreach \name / \y in {1,...,3}           
				\node[input neuron] (I-\name) at (\y + \vert, 0) {$\frac{\partial L}{\partial x_ \y}$};
				
				\foreach \name / \y in {1,...,2}         
				\path[yshift=0.5cm]             
				node[hidden neuron] (H-\name) at (\y + .5 + \vert, - \layersep)  {$\frac{\partial L}{\partial y_ \y}$}; 
				
				\path (H-2) edge[trained] (I-2);    
				\path (H-1) edge[trained] (I-3)  ;    
				
				\begin{scope}
				\node [matrix] [right = \vert of m1] (m2){};
				\node [frozen weight thinn] (w11) at (1.35 + \vert,  1.25 * \layersep) {$w_{1,1}$};    
				\node [frozen weight thinn] [right=0.75pt of w11] (w12){$w_{1,2}$};    
				\node [standard weight thick] [right=0.75pt of w12] (w13){$\tilde{w}_{1,3}$};    
				\node [frozen weight thinn] [below=0.75pt of w11] (w21){$w_{2,1}$};    
				\node [standard weight thick] [right=0.75pt of w21] (w22){$\tilde{w}_{2,2}$};    
				\node [frozen weight thinn] [right=0.75pt of w22] (w23){$w_{2,3}$};
				\end{scope}
				
				\node[annot,right of=hl1, node distance=\vert cm] (hl2) { $\overline{\textbf{\underline{BackProp FreezeNet}}}$};     
				\node[annot,right of=aa1, node distance=\vert cm] (aa2) {\footnotesize Update and store unfrozen weights only};     
				\node[annot,right of=ab1, node distance=\vert cm] (ab2) {\footnotesize Freeze gradient values};

				\foreach \name / \y in {1,...,3}            
				\node[input neuron] (I-\name) at (\y + 2*\vert, 0) {$\frac{\partial L}{\partial x_ \y}$};
				
				\foreach \name / \y in {1,...,2}         
				\path[yshift=0.5cm]             
				node[hidden neuron] (H-\name) at (\y + .5 + 2*\vert, - \layersep)  {$\frac{\partial L}{\partial y_ \y}$};
				
				\path (H-1) edge[frozen] (I-1)  ;    
				\path (H-1) edge[frozen] (I-2)  ;    
				\path (H-1) edge[trained] (I-3)  ;    
				\path (H-2) edge[frozen] (I-1)  ;    
				\path (H-2) edge[trained] (I-2)  ;    
				\path (H-2) edge[frozen] (I-3)  ; 
				
				\begin{scope}
				\node [matrix] [right = \vert of m2] (m3){};
				\node [frozen weight thick] (w11) at (1.35 + 2*\vert,  1.25 * \layersep) {$\tilde{w}_{1,1}$};    
				\node [frozen weight thick] [right=0.75pt of w11] (w12){$\tilde{w}_{1,2}$};    
				\node [standard weight thick] [right=0.75pt of w12] (w13){$\tilde{w}_{1,3}$};    
				\node [frozen weight thick] [below=0.75pt of w11] (w21){$\tilde{w}_{2,1}$};    
				\node [standard weight thick] [right=0.75pt of w21] (w22){$\tilde{w}_{2,2}$};    
				\node [frozen weight thick] [right=0.75pt of w22] (w23){$\tilde{w}_{2,3}$}; 
				\end{scope}
				
				\node[annot,right of=hl2, node distance=\vert cm] (hl3) {$\overline{\textbf{\underline{BackProp Standard NN}}}$};     
				\node[annot,right of=aa2, node distance=\vert cm] (aa3) {\footnotesize Update and store all weights};     
				\node[annot,right of=ab2, node distance=\vert cm] (ab3) {\footnotesize Use all gradient values};
				
				\end{tikzpicture} 
	\end{center}
	
	\caption{\label{fig:freezing}Graphical illustration of FreezeNet and comparison
		with a standard neural network (NN) for a fully connected
		NN with neurons $x_{1},x_{2},x_{3}$ and $y_{1},y_{2}$, and corresponding
		weights and gradient values. Best viewed in colour. }
\end{figure}

SNIP finds the most dominant weights in a neural network with a single
for- and backward pass, performed once before training starts and immediately prunes the other, less important weights. Hence
it is a one-shot pruning method, applied before training. By one-shot pruning we mean pruning in a single step, not iteratively. This leads to sparse gradient computations during training. But if too many
parameters are pruned, SNIP networks are not able to train well due
to a weak flow of the gradient through the network \cite{wang_2020}.
In this work we use a SNIP related method for finding the most influential
weights in a deep neural network (DNN). We do not follow the common
pruning procedure of setting weights to zero, but keep the remaining
parameters fixed as initialized which we call \textit{freezing},
schematically shown in Figure \ref{fig:freezing}. A proper gradient
flow throughout the network can be ensured with help of the frozen
parameters, even for a small number of trained parameters. The frozen
weights also increase the network's expressiveness, without adding
any gradient computations — compared to pruned networks. All frozen
weights can be stored with a single random seed number. We call
these partly frozen DNNs \emph{FreezeNets}.

\subsection{Contributions of this Paper and Applications}

In this work we introduce FreezeNets, which can be applied to any
baseline neural network. The key contributions of FreezeNets are:
\begin{itemize}
	\item Smaller trainable parameter count than one-shot pruning (SNIP), but
	better results.
	\item Preservation of gradient flow, even for a small number of trained
	parameters.
	\item Efficient way to store frozen weights with a single random seed number.
	\item More efficient training than the baseline architecture since the same
	number of gradients as for pruned networks has to be computed.
\end{itemize}
Theoretically and empirically, we show that a faithful gradient
flow, even for a few trainable parameters, can be preserved
by using frozen weights. Whereas pruning weights eventually leads
to vanishing gradients. By applying weight decay also on the frozen
parameters, we modify FreezeNets to generate sparse networks at the
end of training. For freezing rates on which SNIP performs well, this
modified training generates networks with the same number of non-zero weights as SNIP
while reaching better performances. 

Due to their sparse gradient computations, FreezeNets are perfectly
suitable for applications with a train-inference ratio biased towards
training. Especially for research, where networks are trained for
a long time and often validated exactly once, FreezeNets provide a
good trade-off between reducing training resources and keeping performance.
Other applications for FreezeNets are networks that have to be retrained
many times due to changing data, as online learning or transfer learning.
Since FreezeNets can reduce the number of stored parameters drastically,
they are networks cheap to transfer. This could be of interest for
autonomous vehicle fleets or internet services. For a given hardware,
FreezeNets can be used to increase the size of the largest trainable
network since less storage and computations are needed for applying
gradient descent. 

\section{Related Work}

\subsection{Network Pruning}

Pruning methods are used to reduce the amount of parameters in a network
\cite{han_2015,lecun_1990,mozer_1989}.  At the same time, the pruned
network should perform equally well, or even better, than the reference
network. Speeding up training can be achieved by pruning the network
at the beginning of training \cite{lee_2018} or at early training
steps \cite{frankle_2018,frankle_2019}. There are several approaches
to prune neural networks. One is penalizing non-zero weights \cite{cp_2018,chauvin_1989,hanson_1988}
and thus achieving sparse networks. Nowadays, a more common way is
given by using magnitude based pruning \cite{frankle_2018,frankle_2019,guo_2016,han_2015},
leading to pruning early on in training \cite{frankle_2018,frankle_2019},
on-the-fly during training \cite{guo_2016} or at the end of it \cite{han_2015}.
These pruned networks have to be fine-tuned afterwards. For high
pruning rates, magnitude based pruning works better if this procedure
is done iteratively \cite{frankle_2018,frankle_2019}, therefore leading
to many \textit{train-prune(-retrain)} cycles.  
Pruning can also be achieved by using \emph{neural architecture search} \cite{dong_2019,he_2018} or adding computationally cheap branches to predict sparse locations in feature maps \cite{dong_2017}.
The final pruning strategy we want to present is saliency based pruning. In saliency
based pruning, the significance of weights is measured with the Hessian
of the loss \cite{lecun_1990}, or the sensitivity of the loss with
respect to inclusion/exclusion of each weight \cite{karnin_1990}.
This idea of measuring the effect of inclusion/exclusion of weights
was resumed in \cite{lee_2018}, where a differentiable approximation
of this criterion was introduced, the SNIP method. Since SNIP's pruning step is applicable with a single for- and backward pass one-shot
before training, its computational overload is negligible. 
The GraSP (Gradient Signal Preservation) \cite{wang_2020} method is also
a pruning mechanism, applied one-shot before training. Contrarily
to SNIP, they keep the weights possessing the best gradient flow at
initialization. For high pruning rates, they achieve better results
than SNIP but are outperformed by SNIP for moderate ones.

\emph{Dynamic sparse training} is a strategy to train pruned networks, but give the sparse architecture a chance to change dynamically during training.
Therefore, pruning- and regrowing steps have to be done during the whole training process.
The weights to be regrown are determined by random processes \cite{bellec_2018,mocanu_2018,xie_2019}, their magnitude \cite{wortsman_2019} or saliency \cite{dettmers_2019,ding_2019}. An example of the latter strategy is \emph{Sparse Momentum} \cite{dettmers_2019}, measuring saliencies via exponentially smoothed gradients. \emph{Global Sparse Momentum} \cite{ding_2019} uses a related idea to FreezeNet by not pruning the untrained weights. But the set of trainable weights can change and the untrained weights are not updated via gradient descent, but with a momentum parameter based on earlier updates. 
Whereas FreezeNet freezes weights and uses a fixed architecture, thus needs to gauge the best sparse network for all phases of training. 

In pruning, the untrained parameters are set to $0$ which is not
done for FreezeNets, where these parameters are frozen and used to
increase the descriptive power of the network. This clearly separates
our freezing approach from pruning methods. 

\subsection{CNNs with Random Weights\label{subsec:CNNs-with-Random}}

The idea of fixing randomly initialized weights in Convolutional Neural
Networks (CNNs) was researched in \cite{jarrett_2009}, where the
authors showed that randomly initialized convolutional filters act
orientation selective. In \cite{saxe_2011} it was shown that randomly
initialized CNNs with pooling layers can act inherently frequency
selective. Ramanujan et al. \cite{ramanujan_2019} showed that in
a large randomly initialized base network ResNet$50$ \cite{He2015}
a smaller, untrained subnetwork is hidden that matches the performance
of a ResNet$34$ \cite{He2015} trained on ImageNet \cite{deng_2009}.
Recently, Frankle et al. \cite{frankle_2020} published an investigation
of CNNs with only Batch Normalization \cite{ioffe_2015} parameters
trainable. In contrast to their work, we also train biases and chosen weight parameters
and reach competitive results with FreezeNets.

A follow-up work of the \emph{Lottery Ticket Hypothesis} \cite{frankle_2018}
deals with the question of why iterative magnitude based pruning works
so well \cite{zhou_2019}. Among others, they also investigate resetting
pruned weights to their initial values and keeping them fix. The unpruned
parameters are reset to their initial values as well and trained again.
This train-prune-retrain cycle is continued until the target rate
of fixed parameters is reached. In their experiments they show that
this procedure mostly leads to worse results than standard iterative pruning
and just outperforms it for extremely high pruning rates. 


\section{FreezeNets\label{sec:Theoretical-Background}}


\subsubsection{General Setup}

Let $f_\Theta:\mathbb{R}^{d_{0}}\rightarrow[0,1]^{c}$ be a DNN with parameters
$\Theta\subset\mathbb{R}$, used for an image classification task
with $c$ classes. We assume a train set $(X,Z)$ with images ${X=\{x_{1},\ldots,x_{N}\}\subset\mathbb{R}^{d_{0}}}$
and corresponding labels $Z=\{z_{1},\ldots,z_{N}\}\subset\{0,1,\ldots,c-1\}$,
a test set $(X_{0},Z_{0})$ and a smooth loss function $L$ to be
given. As common for DNNs, the test error is minimized by training
the network with help of the training data via stochastic gradient
based (SGD) optimization \cite{robbins_2007} while preventing the
network to overfit on the training data.

We define the rate $q:=1-p$ as the networks \textit{freezing rate},
where $p$ is the rate of trainable weights. A high freezing rate
corresponds to few trainable parameters and therefore sparse gradients, whereas a low freezing rate corresponds to many trainable
parameters. Freezing is compared to pruning in Section \ref{sec:Experiments}.
For simplicity, a freezing rate $q$ for pruning a network means exactly
that $q\cdot100\%$ of its weights are set to zero. In this work we
split the model's parameters into weights $W$ and biases $B$,
only freeze parts of the weights $W$ and keep all biases trainable.

\subsection{SNIP}

Since pruned networks are constraint on using only parts of their
weights, those weights should be chosen as the most influential ones
for the given task.  Let ${\Theta=W \cup B}$
\footnote{By an abuse of notation, we also use $W$ and $B$ as the vectors
	containing all elements of the set of all weights and biases, respectively.} 
be the network's parameters and $m\in\{0,1\}^{\vert W\vert}$
be a mask that shows if a weight is activated or not. Therefore,
the weights that actually contribute to the network's performance
are given by $m\odot W$. Here $\odot$ denotes the Hadamard product
\cite{chandler_1962}. The trick used in \cite{lee_2018} is to
look at the \textit{saliency score}

\begin{equation}
g:=\left.\frac{\partial L(m\odot W;B,X,Z)}{\partial m}\right|_{m=1}=\frac{\partial L(W;B,X,Z)}{\partial W}\odot W\text{\;},\label{eq:grad_saliency}
\end{equation}
which calculates componentwise the influence of the loss function's change 
by a small variation of the  associated weight's activation.\footnote{To obtain differentiability in equation \eqref{eq:grad_saliency}, the mask is assumed to be continuous, i.e. $m\in\mathbb{R}^{\vert W\vert}$.} If those
changes are big, keeping the corresponding weight is likely to have
a greater effect in minimizing the loss function than keeping a weight
with a small score. The gradient $g$ can be approximated with just
a single forward and backward pass of one training batch before the
beginning of training. 

\subsection{Backpropagation in Neural Networks\label{subsec:Backpropagation-in-Neural}}

To simplify the backpropagation formulas, we will deal with a feed-forward,
fully connected neural network. Similar equations hold for convolutional
layers \cite{He2015}. Let the input of the network be given by $x^{(0)}\in\mathbb{R}^{d_{0}}$,
the weight matrices are given by $W^{(k)}\in\mathbb{R}^{d_{k}\times d_{k-1}}$,
$k\in\{1,\ldots,K\}$ and the forward propagation rules are inductively
defined as
\begin{itemize}
	\item $y^{(k)}:=W^{(k)}x^{(k-1)}+b^{(k)}$ for the layers bias $b^{(k)}\in\mathbb{R}^{d_{k}}$
	,
	\item $x^{(k)}:=\Phi_{(k)}(y^{(k)})$ for the layers non-linearity $\Phi_{(k)}:\mathbb{R}\rightarrow\mathbb{R}$,
	applied component-wise.
\end{itemize}
This leads to the partial derivatives used for the backward pass,
written compactly in vector or matrix form:

\begin{equation}
\begin{aligned}\frac{\partial L}{\partial y^{(k)}} & =\Phi_{(k)}^{\prime}\left(y^{(k)}\right)\odot\frac{\partial L}{\partial x^{(k)}}\;, & \frac{\partial L}{\partial x^{(k)}} & =\left(W^{(k+1)}\right)^{T}\cdot\frac{\partial L}{\partial y^{(k+1)}}\;,\\
\frac{\partial L}{\partial W^{(k)}} & =\frac{\partial L}{\partial y^{(k)}}\cdot\left(x^{(k-1)}\right)^{T}\;, & \frac{\partial L}{\partial b^{(k)}} & =\frac{\partial L}{\partial y^{(k)}}
\end{aligned}
\;.\label{eq:backprop_combined}
\end{equation}
Here, we define $W^{(K+1)}:=\text{id}\in\mathbb{R}^{d_{K}\times d_{K}}$
and $\frac{\partial L}{\partial y^{(K+1)}}:=\frac{\partial L}{\partial x^{(K)}}$.
For sparse weight matrices $W^{(k+1)}$, equations (\ref{eq:backprop_combined})
can lead to small $\frac{\partial L}{\partial x^{(k)}}$ and consequently
small weight gradients $\frac{\partial L}{\partial W^{(k)}}$. In
the extreme case of $\frac{\partial L}{\partial y^{(k)}}=0$ for a
layer $k$, 
 all overlying layers will have $\frac{\partial L}{\partial W^{(l)}}=0$,
$l\leq k$. Overcoming the gradient's drying up for sparse weight
matrices in the backward pass motivated us to freeze weights instead
of pruning them.

\subsection{FreezeNet}

\begin{algorithm}[tb]
	\caption{FreezeNet \label{alg:RSNIP}}
	
	\begin{algorithmic}[1]
		\Require{Freezing rate $q$, initial parametrization $\Theta_0=W_0 \cup B_0$, corresponding network $f_{\Theta_0}$, loss function $L$} 
		\State{Compute saliency score $g \in \mathbb{R}^{\vert W_0 \vert}$ for one training batch, according to equation (\ref{eq:grad_saliency})}
		\State{Define freezing mask $m \in \mathbb{R}^{\vert W_0 \vert}$}
		\State{Use freezing threshold $\varepsilon$ as the $\lfloor (1 - q) \cdot \vert W_0 \vert \rfloor$-highest magnitude of $g$}
		\State{Set $m_k = 0$ if $\vert g_k \vert < \varepsilon$ else $m_k = 1$}
		\State{Start training with forward propagation as usual but backpropagate gradient ${m \odot\frac{\partial L}{\partial W_0}}$} for weights and $\frac{\partial L}{\partial B_0}$ for biases
	\end{algorithmic}
	
\end{algorithm}
In Algorithm \ref{alg:RSNIP} the FreezeNet method is introduced.
First, the saliency score $g$ is calculated according to equation
(\ref{eq:grad_saliency}). Then, the freezing threshold $\varepsilon$
is defined as the $\lfloor (1 - q) \cdot \vert W_0 \vert \rfloor$-highest
magnitude of $g$. If a saliency score is smaller than the freezing
threshold, the corresponding entry in the freezing mask $m\in\mathbb{R}^{\vert W_{0}\vert}$
is set to $0$. Otherwise, the entry in $m$ is set to $1$. However,
we do not delete the non-chosen parameters as done for SNIP pruning,
but leave them as initialized. This is achieved with the masked gradient.
For computational and storage capacity reasons, it is more efficient
to not calculate the partial derivative for the weights with mask
value $0$, than masking the gradient after its computation. 

The amount of memory needed to store a FreezeNet is the same as for
standard pruning. With the help of pseudo random number generators,
as provided by PyTorch \cite{pytorch} or TensorFlow \cite{tensorflow},
just the seed used for generating the initial parametrization has
to be stored, which is usually an integer and therefore its memory
requirement is neglectable. The used pruning/freezing mask together with the trained weights have to
be saved for both, pruning and FreezeNets. The masks can be stored
efficiently via entropy encoding \cite{duda_2015}.

In this work, we only freeze weights and keep all biases
learnable, as done in the pruning literature \cite{frankle_2018,frankle_2019,lee_2018,wang_2020}. Therefore, we compute the
freezing rate as $q=1-\frac{\Vert m\Vert_{0}}{\vert W\vert}$, where
$m$ is the freezing mask calculated for the network's weights $W$.
Here, the pseudo norm $\Vert\cdot\Vert_{0}$ computes the number of
non-zero elements in $m$. Since we deal with extremely high freezing
rates, $q>0.99$, the bias parameters have an effect on the percentage
of all trained parameters. Thus, we define the real freezing rate
$q_{\beta}=1-\frac{\text{\ensuremath{\Vert}}m\Vert_{0}+\vert B\vert}{\vert W\vert+\vert B\vert}$
and label the $x$-axes in our plots with both rates.

Pruned networks use masked weight tensors $m\odot W$ in the for-
and backward pass. In theory, the number of computations needed
for a pruned network can approximately be reduced by a factor of $q_{\beta}$
in the forward pass. The frozen networks do not decrease the number
of calculations in the forward pass. But without the usage of specialized
soft- and hardware, the number of computations performed by a pruned
network is not reduced, thus frozen and pruned networks have the same
speed in this setting. 

In the backward pass, the weight tensor needed to compute $\frac{\partial L}{\partial x^{(k-1)}}$
is given by $m^{(k)}\odot W^{(k)}$ for a pruned network, according
to the backpropagation equations (\ref{eq:backprop_combined}). Frozen
networks compute $\frac{\partial L}{\partial x^{(k-1)}}$ with a dense
matrix $W^{(k)}$. On the other hand, not all weight gradients are
needed, as only $m^{(k)}\odot\frac{\partial L}{\partial W^{(k)}}$
is required for updating the network's unfrozen weights. Therefore,
the computation time in the backward pass is not reduced drastically
by FreezeNets, although the number of gradients to be stored. Again,
the reduction in memory is approximately given by the rate $q_{\beta}$.
The calculation of $\frac{\partial L}{\partial x^{(k-1)}}$ with a
dense matrix $W^{(k)}$ helps to preserve a faithful gradient throughout
the whole network, even for extremely high freezing rates, as shown
in Section \ref{subsec:Gradient-Flow}. The comparison of training a pruned
and a frozen network is summarized in Table \ref{tab:Comparison-of-baseline}.
\begin{table}[tb]
	\caption{\label{tab:Comparison-of-baseline}Comparison of standard training, pruning before training and a FreezeNet.}
	\begin{tabular}{>{\centering}m{0.12\textwidth}>{\centering}m{0.15\textwidth}>{\centering}m{0.15\textwidth}>{\centering}m{0.15\textwidth}>{\centering}m{0.2\textwidth}>{\centering}m{0.17\textwidth}}
		\toprule
		Method & $\#$ Total Weights & $\#$ Weights to Store & Sparse Gradients & Sparse Tensor Computations  & Faithful Gradient Flow\tabularnewline
		\midrule
		Standard & $D$ & $D$ & \xmark & \xmark & \cmark\tabularnewline
		Pruned & $D\cdot(1-q)$ & $D\cdot(1-q)$ & \cmark & \cmark & \xmark\tabularnewline
		FreezeNet & $D$ & $D\cdot(1-q)$ & \cmark & \xmark & \cmark\tabularnewline
		\bottomrule
	\end{tabular}
\end{table}

\section{Experiments and Discussions\label{sec:Experiments}}

In the following, we present results on the MNIST \cite{lecun_1998}
and CIFAR-$10$/$100$ \cite{krizhevsky_2012} classification tasks
achieved by FreezeNet. Freezing networks is compared with training
sparse networks, exemplified through SNIP \cite{lee_2018}. We further
analyse how freezing weights retains the trainability of networks
with sparse gradient updates by preserving a faithful gradient. We
use three different network architectures, the fully connected LeNet-$300$-$100$
\cite{lecun_1998} along with the CNN's LeNet-5-Caffe \cite{lecun_1998}
and VGG$16$-D \cite{simonyan_2014}. A more detailed description
of the used network architectures can be found in the Supplementary
Material.  Additionally, we show in the Supplementary Material that FreezeNets based on a ResNet$34$ perform well on Tiny ImageNet.

For our experiments we used PyTorch $1.4.0$ \cite{pytorch} and a
single Nvidia GeForce $1080$ti GPU. In order to achieve a fair
comparison regarding hard- and software settings we recreated SNIP.\footnote{Based on the official implementation https://github.com/namhoonlee/snip-public.}
To prevent a complete loss of information flow we randomly flag one
weight trainable per layer if all weights of this layer are frozen
or pruned for both, SNIP and FreezeNet. This adds at most $3$ trainable
parameters for LeNet-$300$-$100$, $4$ for LeNet-$5$-Caffe and
$16$ for {VGG$16$\hbox{-}D}. If not mentioned otherwise, we use
Xavier-normal initializations \cite{xavier_2010} for SNIP and 
FreezeNets and apply weight decay on the trainable parameters only.
Except where indicated, the experiments were run five times with different
random seeds, resulting in different network initializations, data
orders and additionally for CIFAR experiments in different data augmentations.
In our plots we show the mean test accuracy together with one standard
deviation. A split of $9/1$ between training examples and validation
examples is used for early stopping in training. All other hyperparameters
applied in training are listed in the Supplementary Material. 

SGD with momentum \cite{sutskever_2013} is used as optimizer, thus
we provide a learning rate search for FreezeNets in the Supplementary
Material. Because $\lambda=0.1$ works best for almost all freezing
rates, we did not include it in the main body of the text and use
$\lambda=0.1$ with momentum $0.9$ for all presented results. Altogether, we use the same setup as SNIP in \cite{lee_2018} for both, FreezeNets and SNIP pruned networks. 

\subsection{MNIST \label{subsec:MNIST-with-LeNet--} \label{subsec:MNIST-with-LeNet--Caffe}}

\begin{figure}[tb]
	\includegraphics[width=0.5\textwidth]{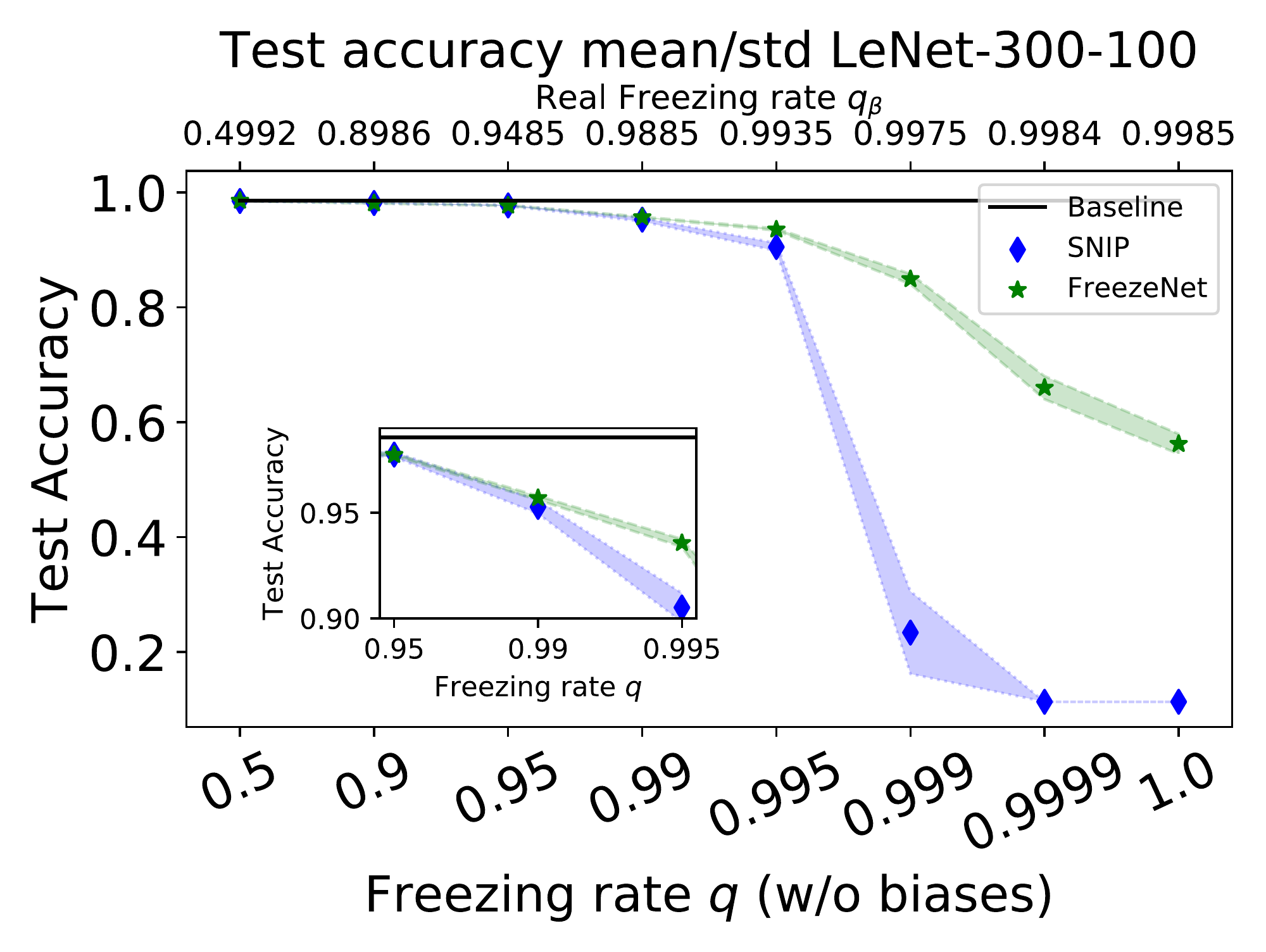}\includegraphics[width=0.5\textwidth]{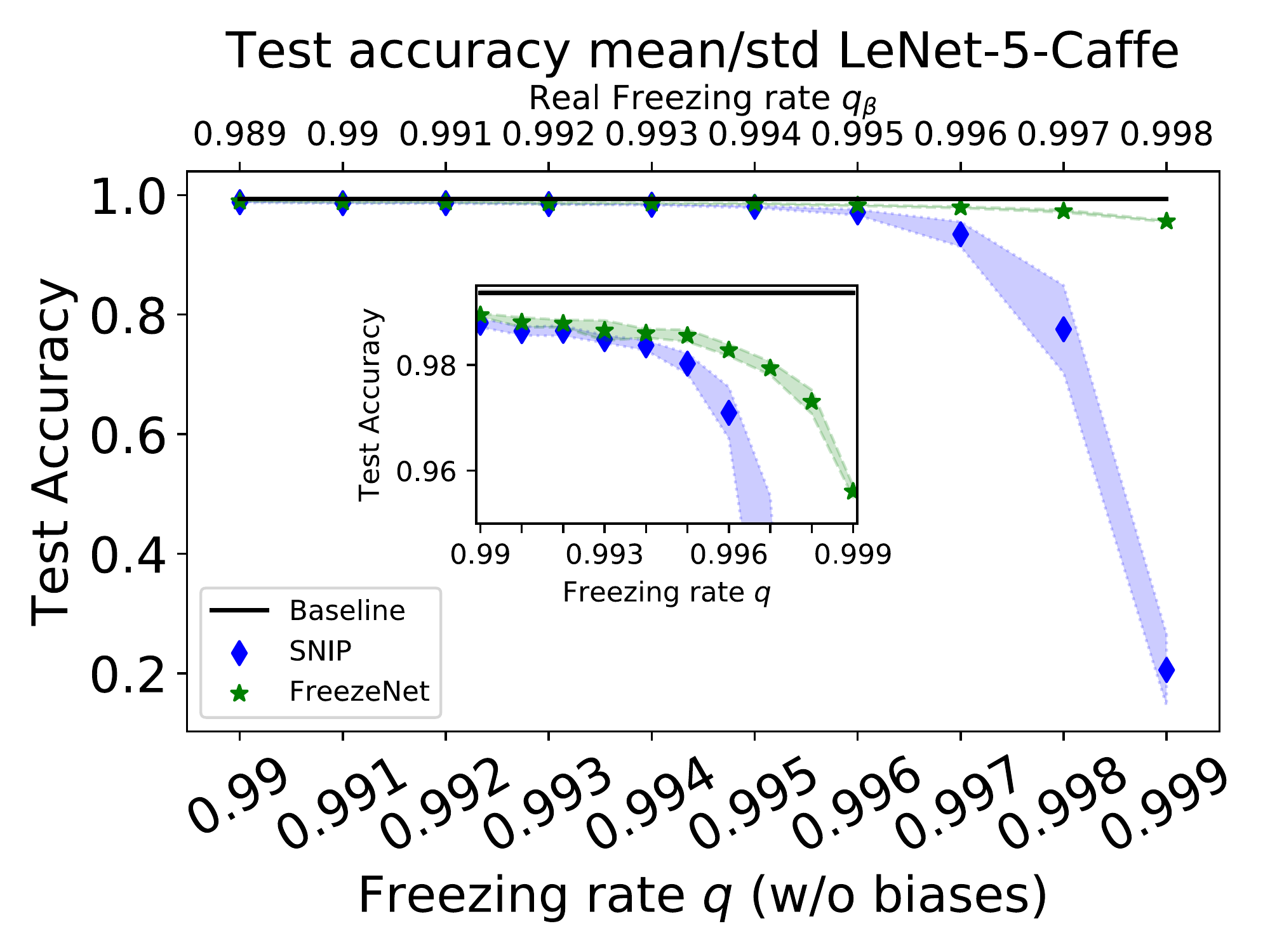}
	
	\caption{\label{fig:lenet_snip_rsnip}Left: Test accuracy for SNIP,
		FreezeNet and the baseline LeNet-$300$-$100$. 
		Right: Test accuracy for SNIP and
		FreezeNet for a LeNet-5-Caffe baseline. The small inserted plots
		are zoomed in versions for both plots. }
\end{figure}

\subsubsection{LeNet-$300$-$100$}

A common baseline to examine pruning mechanisms on fully connected
networks is given by testing the LeNet-$300$-$100$ \cite{lecun_1998}
network on  the MNIST classification task \cite{lecun_1998}, left
part of Figure \ref{fig:lenet_snip_rsnip}.  The trained baseline
architecture yields a mean test accuracy of $98.57\%$. If the freezing
rate is lower than $0.95$, both methods perform equally well and
also match the performance of the baseline. For higher freezing rates,
the advantage of using free, additional parameters can be seen. FreezeNets
also suffer from the loss of trainable weights, but they are able to
compensate it better than SNIP pruned networks do.

\begin{table}[tb]
	\caption{\label{tab:Comparsion-of-FreezeNet}Comparison of FreezeNet, SNIP and the
		LeNet-$5$-Caffe baseline. Results for different \emph{freezing rates} $q$ with corresponding \emph{real freezing rates} $q_\beta$ are displayed. The \emph{network's size} is calculated
		without compression. Thus, all weights are stored as $32$bit floats.
		\emph{Compress. Factor FN} is the compression factor gained
		by FreezeNet for the corresponding freezing rate, calculated via the ratio of the
		\emph{network sizes} of the baseline and the frozen network. }
	
	\begin{tabular}{>{\centering}m{0.12\textwidth}>{\centering}m{0.12\textwidth}>{\centering}m{0.12\textwidth}>{\centering}m{0.14\textwidth}>{\centering}m{0.14\textwidth}>{\centering}m{0.14\textwidth}>{\centering}m{0.15\textwidth}}
		\toprule
		$q$ & $q_{\beta}$ & Network Size & Compress. Factor FN& Test Acc. SNIP & Test Acc. FreezeNet & $\frac{\text{FreezeNet Acc.}}{\text{Baseline Acc.}}$\tabularnewline
		\midrule 
		\multicolumn{2}{c}{$0$ (Baseline)} & $1,683.9$kB & $1$ & \multicolumn{2}{c}{\underline{$99.36\%$}} & $1.000$\tabularnewline
		\midrule 
		$0.9$ & $0.899$ & $170.7$kB & $\bm{9.9}$ & $99.24\%$ & $\bm{99.37\%}$ & $1.000$\tabularnewline
		$0.99$ & $0.989$ & $19.1$kB & $\bm{88.2}$ & $98.80\%$ & $\bm{98.94\%}$ & $0.996$\tabularnewline
		$0.995$ & $0.994$ & $10.7$kB & $\bm{157.4}$ & $98.02\%$ & $\bm{98.55\%}$ & $0.992$\tabularnewline
		$0.999$ & $0.998$ & $3.9$kB & $\bm{431.8}$ & $20.57\%$ & $\bm{95.61\%}$ & $0.962$\tabularnewline
		\bottomrule
	\end{tabular}
\end{table}

\subsubsection{LeNet-$5$-Caffe}

For moderate freezing rates $q\in[0.5,0.95]$, again FreezeNet and
SNIP reach equal results and match the baseline's performance as shown
in the Supplementary Material.   In the right part of Figure
\ref{fig:lenet_snip_rsnip}, we show the progression of SNIP and FreezeNet
for more extreme freezing rates $q\in\{0.99,0.991,\ldots,0.999\}$.
Until $q=0.994$ SNIP and FreezeNet perform almost equally, however
FreezeNet reaches slightly better results.  For higher freezing
rates, SNIP's performance drops steeply whereas FreezeNet is able
to slow this drop. As Table \ref{tab:Comparsion-of-FreezeNet}
and Figure \ref{fig:lenet_snip_rsnip} show, a FreezeNet saves parameters
with respect to both, the baseline architecture and a SNIP pruned
network. 
\begin{figure}[tb]
	\includegraphics[width=0.5\textwidth]{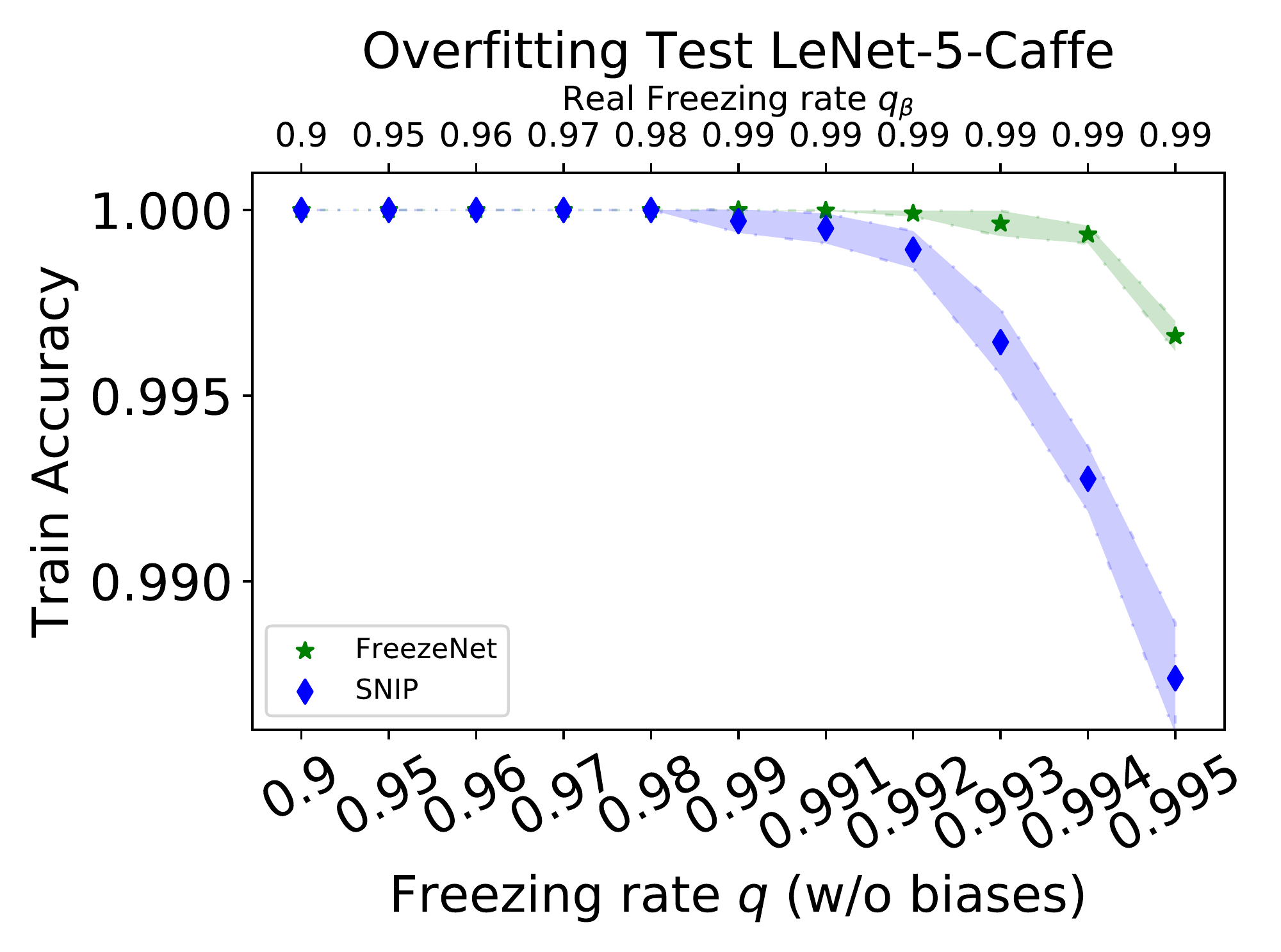}\includegraphics[width=0.5\textwidth]{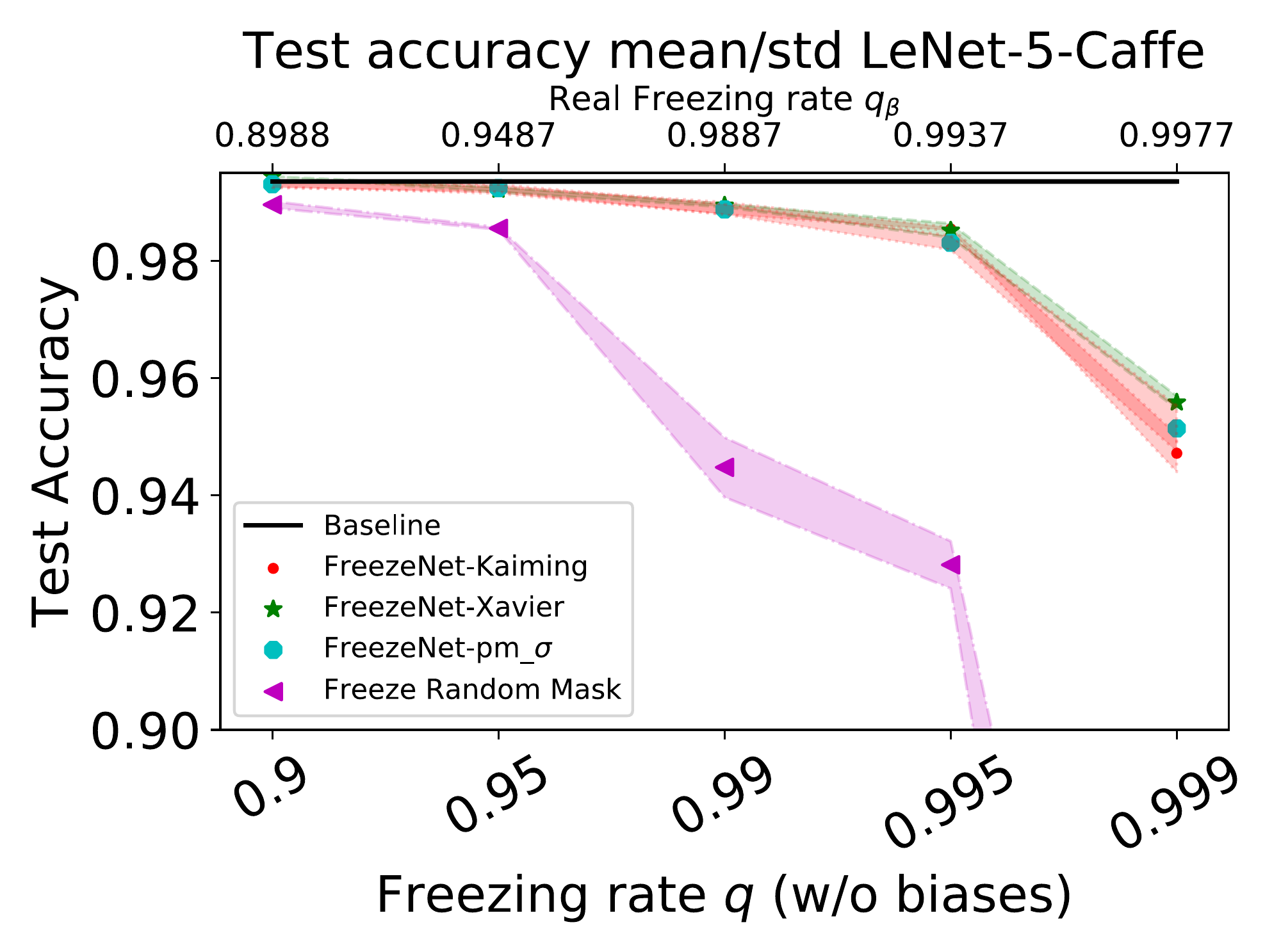}\caption{\label{fig:capcity_init}Left: Final training accuracies
		for FreezeNet and SNIP, both trained without weight decay. Right: Different initializations for FreezeNets together with a Xavier-normal
		initialized
		FreezeNet with randomly generated freezing mask. Both plots are reported for the MNIST classification
		task with a LeNet-$5$-Caffe baseline architecture.}
	
\end{figure}


In order to overfit the training data maximally, we change the training
setup by training the networks without the usage of weight decay and
early stopping. In the left part of Figure \ref{fig:capcity_init},
the training accuracies of FreezeNet and SNIP are reported for the
last training epoch. Unsurprisingly, too many frozen parameters reduce
the model's capacity, as the model is not able to perfectly memorize
the training data for rates higher than $q_{\ast}=0.992$. On the
other hand, FreezeNets increase the networks capacity compared to
SNIP if the same, high freezing rate is used.

\subsection{Testing FreezeNets for CIFAR-$10$/$100$ on VGG$16$-D\label{subsec:Testing-GreeNets-for}}

\begin{table}[tb]
	\begin{centering}
		\caption{\label{tab:cifar_results}Comparison of results for the CIFAR-$10$/$100$
			tasks with a VGG$16$-D baseline. }
		\par\end{centering}
	\centering{}%
	\begin{tabular}{>{\raggedright}m{0.2\textwidth}>{\centering}m{0.2\textwidth}>{\centering}m{0.15\textwidth}>{\centering}m{0.2\textwidth}>{\centering}m{0.2\textwidth}}
		\toprule
		&  &  & \multicolumn{1}{c}{\bf CIFAR-$\bm{10}$} & \multicolumn{1}{c}{\bf CIFAR-$\bm{100}$}\tabularnewline
		\midrule
		Method & Freezing Rate & Trained Parameters &Mean $\pm$ Std& Mean $\pm$ Std\tabularnewline
		\midrule 
		Baseline & $0$ & $15.3$mio & \multicolumn{1}{c}{$\bm{93.0\pm0.1\%}$} & \multicolumn{1}{c}{$\bm{71.6\pm0.6}\%$}\tabularnewline
		\midrule 
		\multirow{4}{0.2\textwidth}{SNIP} & $0.9$ & $1.5$mio & $92.9\pm0.1\%$ & $53.9\pm1.7\%$\tabularnewline
		& $0.95$ & $780$k & $92.5\pm0.1\%$ & $48.6\pm6.6\%$\tabularnewline
		& $0.99$ & $169$k & $10.0\pm0.0\%$ & $1.0\pm0.0\%$\tabularnewline
		& $0.995$ & $92$k & $10.0\pm0.0\%$ & $1.0\pm0.0\%$\tabularnewline
		\midrule 
		\multirow{4}{0.2\textwidth}{FreezeNet} & $0.9$ & $1.5$mio & $92.2\pm0.1\%$ & $\bm{70.7\pm0.3}\%$\tabularnewline
		& $0.95$ & $780$k & $91.7\pm0.1\%$ & $\bm{69.0\pm0.2}\%$\tabularnewline
		& $0.99$ & $169$k & $\bm{88.6\pm0.1\%}$ & $\bm{59.8\pm0.3}\%$\tabularnewline
		& $0.995$ & $92$k & $\bm{86.0\pm0.1\%}$ & $\bm{53.4\pm0.1}\%$\tabularnewline
		\midrule 
		\multirow{4}{0.2\textwidth}{FreezeNet-WD} & $0.9$ & $1.5$mio & $\bm{93.2\pm0.2\%}$ & $53.1\pm1.8\%$\tabularnewline
		& $0.95$ & $780$k & $\bm{92.8\pm0.2\%}$ & $44.5\pm5.4\%$\tabularnewline
		& $0.99$ & $169$k & $76.1\pm1.0\%$ & $13.1\pm1.8\%$\tabularnewline
		& $0.995$ & $92$k & $74.6\pm1.1\%$ & $11.9\pm1.4\%$\tabularnewline
		\bottomrule
	\end{tabular}
\end{table}
To test FreezeNets on bigger architectures, we use the VGG$16$-D
architecture \cite{simonyan_2014} and the CIFAR-$10$/$100$ datasets.
Now, weight decay is applied to all parameters, including
the frozen ones, denoted with FreezeNet-WD. As before, weight decay
is also used on the unfrozen parameters only, which we again call
FreezeNet. We follow the settings in \cite{lee_2018} and exchange
Dropout layers with Batch Normalization \cite{ioffe_2015} layers.
Including the Batch Normalization parameters, the VGG$16$-D network
consists of $15.3$ million parameters in total. We train all Batch
Normalization parameters and omit them in the freezing rate $q$.
Additionally, we augment the training data by random horizontal flipping
and translations up to $4$ pixels. For CIFAR-$100$ we report results
for networks initialized with a Kaiming-uniform initialization \cite{He2015}.
The results are summarized in Table \ref{tab:cifar_results}.

\subsubsection{CIFAR-$10$}

If more parameters are trainable, $q\leq0.95$, SNIP performs slightly
worse than the baseline but  better than FreezeNet. However, using
frozen weights can achieve similar results as the baseline architecture
while outperforming SNIP if weight decay is applied to them as well,
as shown with FreezeNet-WD. Applying weight decay also on the frozen
parameters solely shrinks them to zero. For all occasions where FreezeNet-WD
reaches the best results, the frozen weights can safely be pruned at the early stopping time, as they are all shrunk to zero at this
point in training. For these freezing rates, FreezeNet-WD can be considered
as a pruning mechanism outperforming SNIP without adding any gradient
computations For higher freezing rates $q\geq0.99$, FreezeNet still
reaches reasonable results whereas FreezeNet-WD massively drops performance and SNIP even results in random guessing.

\subsubsection{CIFAR-$100$}

CIFAR-$100$ is more complex to solve than CIFAR-$10$. As the right
part of Table \ref{tab:cifar_results} shows, SNIP is outperformed
by FreezeNet for all freezing rates. Frozen parameters seem to be
even more helpful for a sophisticated task. For CIFAR-$100$, more
complex information flows backwards during training, compared to CIFAR-$10$.
Thus, using dense weight matrices in the backward pass helps to provide
enough information for the gradients to train successfully. Additionally
we hypothesize, that random features generated by frozen parameters
can help to improve the network's performance, as more and often closely
related classes have to be distinguished. Using small, randomly generated
differences between data samples of different, but consimilar classes
may help to separate them.

\subsubsection{Discussion}

Deleting the frozen weights
reduces the network's capacity — as shown for the MNIST task, Figure
\ref{fig:capcity_init} left. But for small freezing rates, the pruned network still has enough capacity in
the forward- and backward propagation. In these cases, the pruned network has
a higher generalization capability than the FreezeNet, according to
the bias-variance trade-off \cite{german_1992}. Continuously decreasing
the network's capacity during training, instead of one-shot, seems
to improve the generalization capacity even more, as done with FreezeNet-WD.
But for higher freezing rates, unshrunken and frozen parameters improve
the performance significantly. For these rates, FreezeNet is still
able to learn throughout the whole training process. Whereas FreezeNet-WD
reaches a point in training, where the frozen weights are almost zero.
Therefore, the gradient does not flow properly through the network,
since the pruned SNIP network has zero gradient flow for these rates, Figure \ref{fig:Relative-average-gradient} left.
This change of FreezeNet-WD's gradient's behaviour is shown in Figure
\ref{fig:Relative-average-gradient} right for $q=0.99$. It should
be mentioned that in these cases, FreezeNet-WD will have an early
stopping time before all frozen weights are shrunk to zero and FreezeNet-WD
can not be pruned without loss in performance.

\subsection{Gradient Flow\label{subsec:Gradient-Flow}}

\begin{figure}[tb]
	\includegraphics[width=0.5\textwidth]{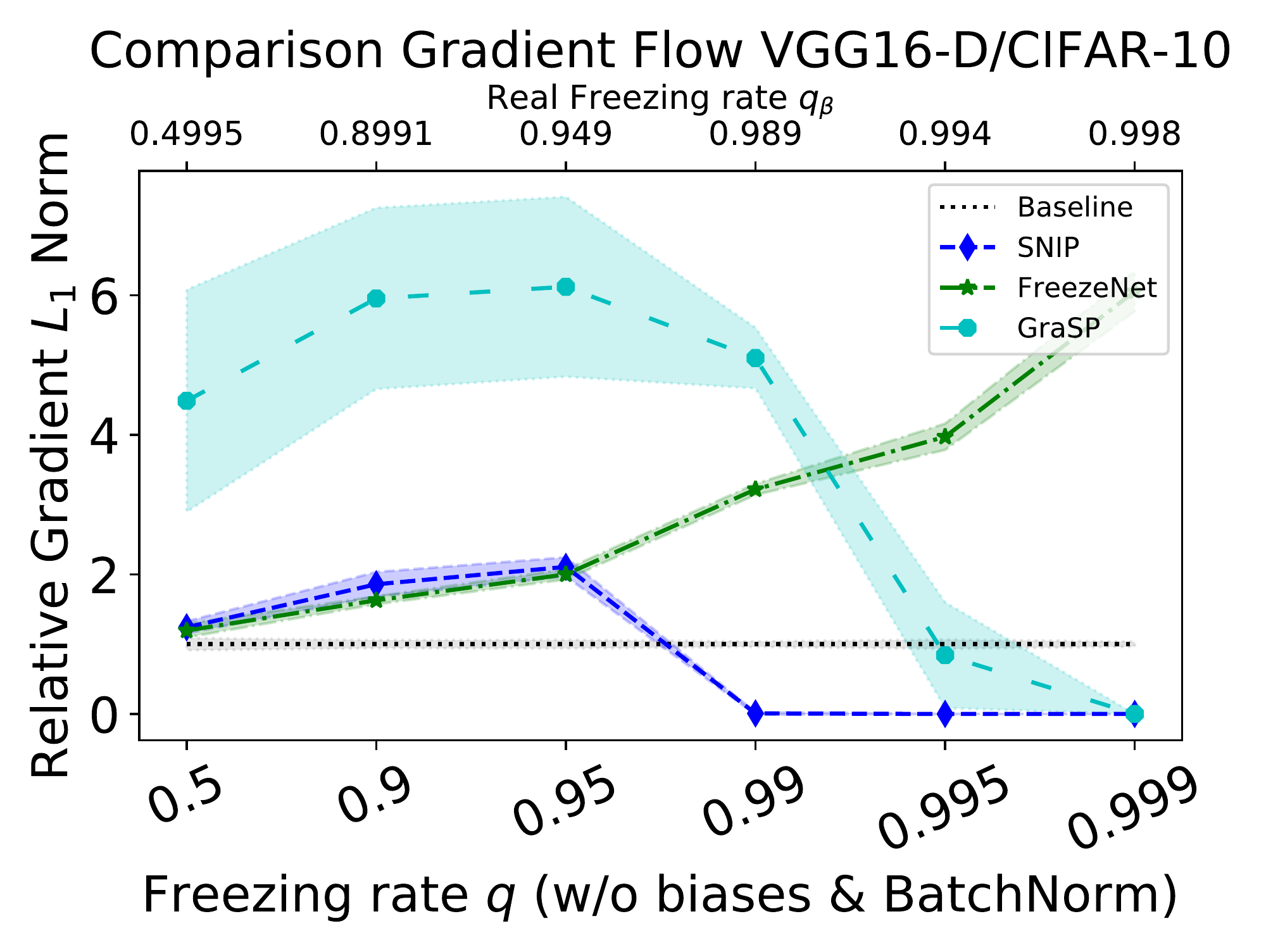}\includegraphics[width=0.5\textwidth]{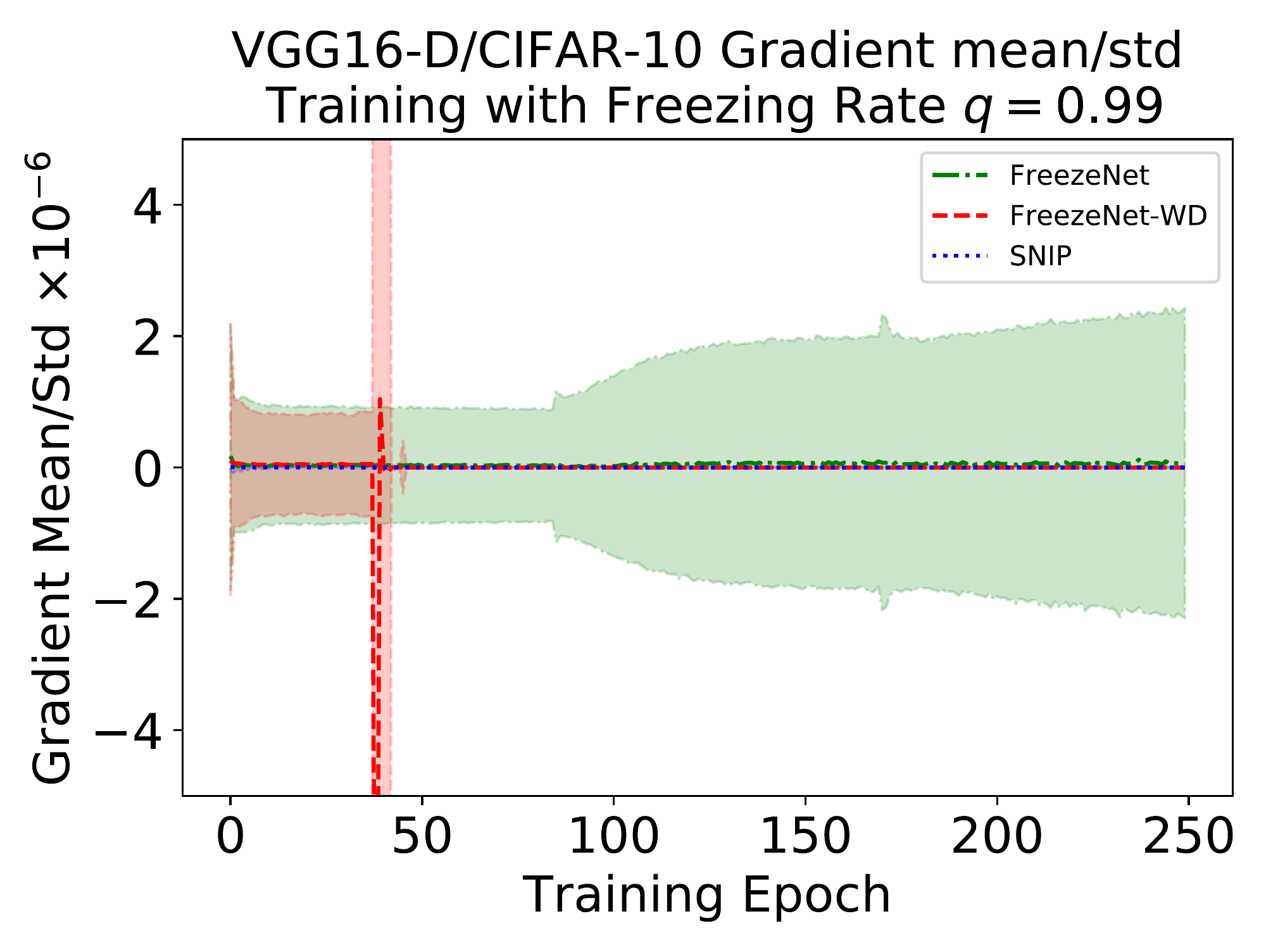}
	
	\caption{\label{fig:Relative-average-gradient}Left: Shows the relative gradient
		norm for FreezeNet, SNIP and GraSP networks with respect to the VGG$16$-D
		baseline network on the CIFAR-$10$ dataset. Right: Gradient mean
		and std, computed over the training data, recorded for one training
		run with a VGG$16$-D architecture on the CIFAR-$10$ task for a freezing
		rate $q=0.99$.}
\end{figure}
As theoretically discussed in Section \ref{subsec:Backpropagation-in-Neural},
FreezeNets help to preserve a strong gradient, even for high freezing
rates. To check this, we also pruned networks with the GraSP criterion
\cite{wang_2020} to compare FreezeNets with pruned networks generated
to preserve the gradient flow. A detailed description of the GraSP
criterion can be found in the Supplementary Material. For this test,
$10$ different networks were initialized for every freezing rate
and three copies of each network were frozen (FreezeNet) or pruned
(SNIP and GraSP), respectively. The $L_{1}$ norm of the gradient,
accumulated over the whole training set, is divided by the number
of trainable parameters. As reference, the mean norm of the baseline
VGG$16$-D's gradient is measured as well. These gradient norms, computed
for CIFAR-$10$, are compared in the left part of Figure \ref{fig:Relative-average-gradient}.
For smaller freezing rates, all three methods have bigger gradient
values than the baseline, on average. For rates $q\geq0.95$, decreasing
the number of trainable parameters leads to a reduced gradient flow
for the pruned networks. Even if the pruning mask is chosen to guarantee
the highest possible gradient flow, as approximately done by GraSP.
Finally, the gradient vanishes,
since the weight tensors become sparse for high pruning rates, as already discussed in Section \ref{subsec:Backpropagation-in-Neural}. FreezeNet's
gradient on the other hand is not hindered since its weight tensors
are dense. The saliency score (\ref{eq:grad_saliency}) is biased
towards choosing weights with a high partial derivative. Therefore,
FreezeNet's non-zero gradient values even become larger as the number of trainable
parameters decreases. For high freezing rates, FreezeNet's gradient
is able to flow through the network during the whole training process,
whereas SNIP's gradient remains zero all the time — right part of
Figure \ref{fig:Relative-average-gradient}. The right part of Figure \ref{fig:Relative-average-gradient} also shows the mutation of FreezeNet-WD's gradient flow during training. First, FreezeNet-WD has similar gradients as FreezeNet since the frozen weights are still big enough.
The red peak indicates the point where too many frozen weights are shrunken close to zero, leading to temporarily chaotic gradients and resulting in zero gradient flow.

\subsection{Comparison to Pruning Methods}

Especially for extreme freezing rates, we see that FreezeNets perform
remarkably better than SNIP, which often degenerates to random guessing.
In Table \ref{tab:Comparison-caffe}, we compare our result for LeNet-$5$-Caffe
with \emph{Sparse-Momentum} \cite{dettmers_2019}, SNIP, GraSP and
three other pruning methods \emph{Connection-Learning} \cite{han_2015},
\emph{Dynamic-Network-Surgery} \cite{guo_2016} and \emph{Learning-Compression} \cite{cp_2018}.
Up to now, all results are reported without any change in hyperparameters.
To compare FreezeNet with other pruning methods, we change the training
setup slightly by using a split of $19/1$ for train and validation
images for FreezeNet, but keep the remaining hyperparameters fixed.
We also recreated results for GraSP \cite{wang_2020}. The training
setup and the probed hyperparameters for GraSP can be found in the Supplementary
Material. All other results are reported from the corresponding papers.
As shown in Table \ref{tab:Comparison-caffe}, the highest accuracy
of $99.2\%$ is achieved by the methods \emph{Connection-Learning}
and \emph{Sparse-Momentum}. With an accuracy of $99.1\%$ our FreezeNet
algorithm performs only slightly worse, however Connection-Learning
trains $8.3\%$ of its weights — whereas FreezeNet achieves $99.37\%$
accuracy with $10\%$ trained weights, see Table \ref{tab:Comparsion-of-FreezeNet}.
Sparse-Momentum trains with sparse weights, but updates the gradients
of all weights during training and redistributes the learnable weights
after each epoch. Thus, their training procedure does neither provide
sparse gradient computations nor one-shot pruning and is hence more
expensive than FreezeNet. Apart from that, FreezeNet achieves similar
results to \emph{Dynamic-Network-Surgery} and better results than
\emph{Learning-Compression}, GraSP and SNIP, while not adding any
training costs over GraSP and SNIP and even reducing them for \emph{Dynamic-Network-Surgery}
and \emph{Learning-Compression}. \begin{table}[tb] 
	\caption{\label{tab:Comparison-caffe}Comparison of different pruning methods with FreezeNet on LeNet-5-Caffe.}
	\begin{tabular}{>{\centering}m{0.375\textwidth}>{\centering}m{0.15\textwidth}>{\centering}m{0.15\textwidth}>{\centering}m{0.15\textwidth}>{\centering}m{0.125\textwidth}} 
		\toprule
		Method & Sparse Gradients in Training & Additional Hyperparameters & Percent of trainable parameters & Test Accuracy
		\tabularnewline 
		\midrule  
		Baseline \cite{lecun_1998} & $-$ & $-$ & $100\%$ & $99.4\%$
		\tabularnewline 
		\midrule  
		SNIP \cite{lee_2018} & \cmark & \xmark & $1.0\%$ & $98.9\%$
		\tabularnewline 
		
		GraSP \cite{wang_2020} & \cmark & \xmark & $1.0\%$ & $98.9\%$
		\tabularnewline 
		
		Connection-Learning \cite{han_2015} & \xmark & \xmark & $8.3\%$ & $99.2\%$
		\tabularnewline
		
		Dynamic-Network-Surgery \cite{guo_2016} & \xmark & \xmark & $0.9\%$ & $99.1\%$
		\tabularnewline
		
		Learning-Compression \cite{cp_2018} & \xmark & \cmark & $1.0\%$ & $98.9\%$
		\tabularnewline
		
		Sparse-Momentum \cite{dettmers_2019} & \xmark & \cmark & $1.0\%$ & $99.2\%$
		\tabularnewline
		
		FreezeNet (ours) & \cmark & \xmark & $1.0\%$ & $99.1\%$
		\tabularnewline 
		\bottomrule
		
	\end{tabular} 
\end{table}

\subsection{Further Investigations\label{sec:Discussions}}

The right part of Figure \ref{fig:capcity_init} shows on the one
hand, that FreezeNet reaches better and more stable results than freezing
networks with a randomly generated freezing mask. This accentuates
the importance of choosing the freezing mask consciously,  for FreezeNets done
with the saliency score (\ref{eq:grad_saliency}).

On the other hand, different variance scaling initialization schemes
are compared for FreezeNets in the right part of Figure \ref{fig:capcity_init}.
Those initializations help to obtain a satisfying gradient flow at the beginning
of the training \cite{xavier_2010,He2015}. Results for the Xavier-normal initialization \cite{xavier_2010}, the Kaiming-uniform \cite{He2015}
and the ${\tt pm}_\sigma$-initialization are shown. All of these
initializations lead to approximately the same results. Considering
all freezing rates, the Xavier-initialization yields the best results.
The ${\tt pm}_\sigma$-initialization is a variance scaling initialization,
using zero mean and a variance of ${\sigma^{2}=\frac{2}{fan_{in}+fan_{out}}}$,
layerwise. All weights are set to $+\sigma$ with probability $\frac{1}{2}$
and $-\sigma$ otherwise. Using the ${\tt pm}_\sigma$-initialization
shows, that even the simplest variance scaling method leads to good
results for FreezeNets.

In the Supplementary Material we exhibit that FreezeNets are robust
against reinitializations of their weights after the freezing mask
is computed and before the actual training starts. The probability
distribution can even be changed between initialization and reinitialization
while still leading to the same performance.

\section{Conclusions}

With FreezeNet we have introduced a pruning related mechanism that
is able to reduce the number of trained parameters in a neural network
significantly while preserving a high performance. FreezeNets match
state-of-the-art pruning algorithms without using their sophisticated
and costly training methods, as Table \ref{tab:Comparison-caffe}
demonstrates.  We showed that frozen parameters help to overcome
the vanishing gradient occurring in the training of sparse neural
networks by preserving a strong gradient signal. They also enhance
the expressiveness of a network with few trainable parameters, especially
for more complex tasks. With the help of frozen weights, the number
of trained parameters can be reduced compared to the related pruning
method SNIP. This saves storage space and thus reduces transfer costs
for trained networks. For smaller freezing rates, it might be better
to weaken the frozen parameters' influence, for example by applying
weight decay to them. Advantageously, using weight decay on frozen
weights contracts them to zero, leading to sparse neural networks.
But for high freezing rates, weight decay in its basic form might not be the best regularization
mechanism to apply to FreezeNets, since only shrinking the frozen
parameters robs them of a big part of their expressiveness in the
forward and backward pass.

\bibliographystyle{./bib/splncs04}
\bibliography{./bib/ref}

\newpage
\setcounter{footnote}{0} 
\renewcommand{\thefootnote}{S.\arabic{footnote}}
\renewcommand{\thepage}{S\arabic{page}}  
\renewcommand{\thesection}{S\arabic{section}}
\renewcommand{\thetable}{S\arabic{table}}   
\renewcommand{\thefigure}{S\arabic{figure}}
\renewcommand{\thealgorithm}{S\arabic{algorithm}}
\renewcommand{\theequation}{S.\arabic{equation}}

\pagenumbering{roman} 

\setcounter{section}{0}
\setcounter{page}{1}
\setcounter{figure}{0}
\setcounter{algorithm}{0}
\setcounter{equation}{0}


\makeatother

\FloatBarrier

\section{Supplementary Material for FreezeNet: Full Performance by Reduced Storage Costs}

\subsection{Training Hyperparameters\label{subsec:Experimental-Setup} for MNIST and CIFAR-$10$}

Our results in the main body of the text are benchmarked against SNIP \cite{lee_2018}, since
we share its feature selection and it is also a one-shot method, applied
before training. For the comparison we use their training schedule
and do not tune any hyperparameters. They take a batch size of $100$
(MNIST) or $128$ (CIFAR) and SGD with momentum parameter $0.9$ \cite{sutskever_2013}
as optimizer. The initial learning rate is set to $0.1$ and divided
by ten at every $25$k/$30$k optimization steps for MNIST and CIFAR,
respectively. For regularization, weight decay with coefficient $5\cdot10^{-4}$
is applied. The overall number of training epochs is $250$. 

\subsection{Tiny ImageNet Experiment}
\begin{figure}[tb]
	\centering
	\includegraphics[width=0.5\textwidth]{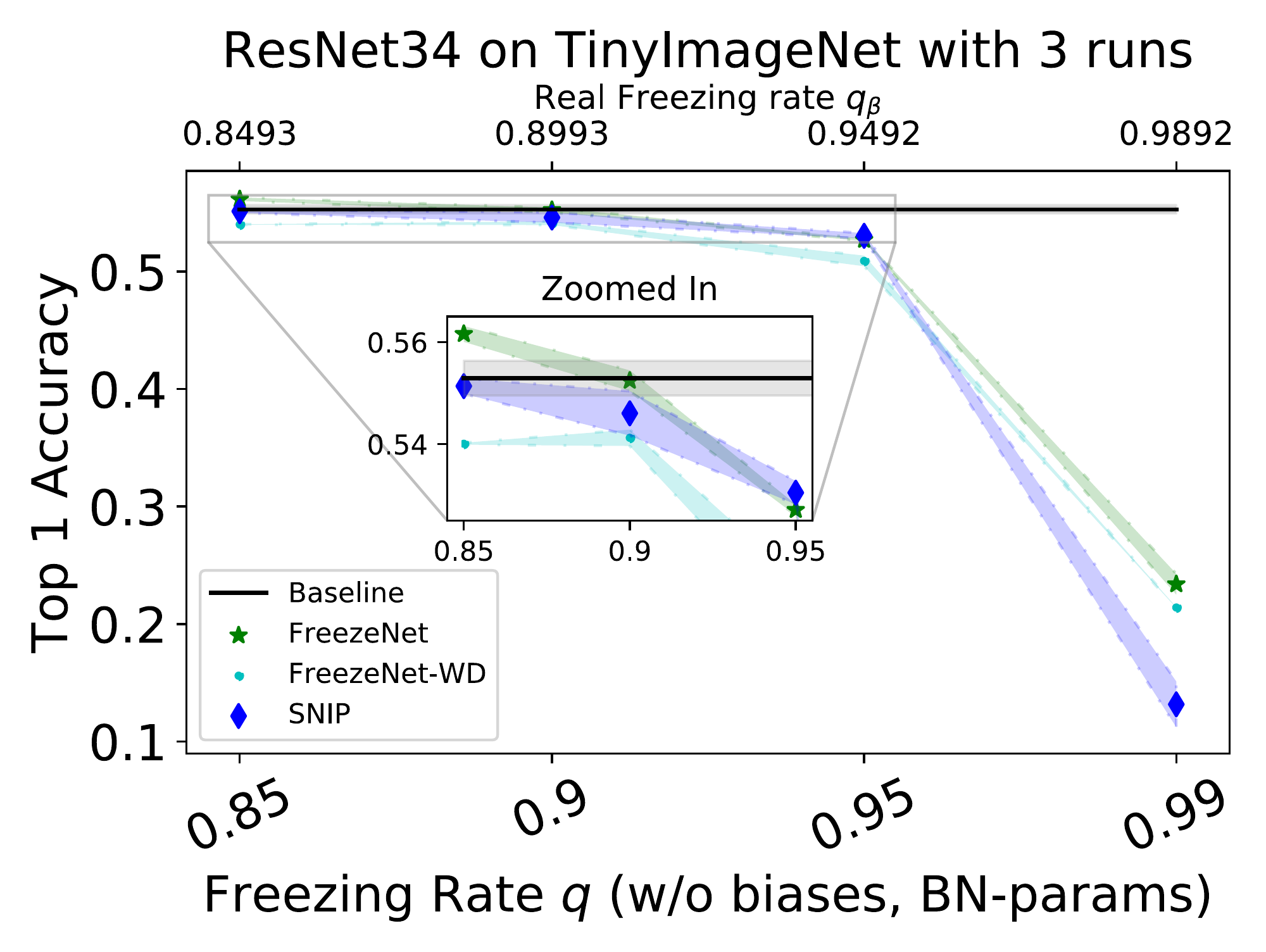}
	
	\caption{\label{fig:tiny}Comparison of FreezeNet, FreezeNet-WD and SNIP for a ResNet$34$ trained on Tiny ImageNet.}
\end{figure}

In addition to the MNIST and CIFAR experiments, we also tested FreezeNet on a ResNet$34$ \cite{He2015} on the Tiny ImageNet classification task \footnote{https://tiny-imagenet.herokuapp.com/ with training schedule adapted from https://github.com/ganguli-lab/Synaptic-Flow.}.
The ResNet$34$ architecture is shown in Table \ref{tab:ResNet}.
The Tiny ImageNet dataset consists of $64\times64$ RGB images with $200$ different classes. Those classes have $500$ training images, $50$ validation images and $50$ test images, each.  
For training, we use the same data augmentation as for CIFAR, i.e. random horizontal flipping and translations up to $4$ pixels. We train the network for $100$ epochs with an initial learning 
rate of $0.01$. The learning rate is decayed by a factor $10$ after epoch $30$, $60$ and $80$. As optimizer, again SGD with momentum parameter $0.9$ is used. Also, weight decay with coefficient $5\cdot10^{-4}$
is applied. The networks are initialized with a Kaiming-uniform initialization.

In Figure \ref{fig:tiny}, the results for FreezeNet, FreezeNet-WD and SNIP are shown. The mean and standard deviation for every freezing rate and method are each calculated for three runs with different random seeds.
We see, that FreezeNet can be used to train a ResNets successfully. FreezeNet has the same accuracy than the baseline method while training only $10\%$ of its weights.
For all rates, FreezeNet achieves better or equal results than SNIP. Also, FreezeNet's results are more stable, shown by the slimmer standard deviation bands. Contrarily to the CIFAR-$10$ experiment, FreezeNet-WD 
shows worse results than SNIP for lower freezing rates. Reaching good results  with a FreezeNet on ResNet$34$ needs a higher rate of trainable parameters than on VGG$16$, see Table \ref{tab:cifar_results}. 
\subsection{GraSP}

The GraSP criterion \cite{wang_2020} is based on the idea of preserving,
or even increasing, the pruned network's gradient flow compared to
the unpruned counterpart. Thus, \begin{equation}
\Delta L(W):=\left\Vert \frac{\partial L}{\partial W}\right\Vert _{2}^{2}
\end{equation}is tried to be maximized with the masked weights $m\odot W$. For
that purpose, \begin{equation}
\Delta L(m\odot W)\approx\Delta L(W)-W^{T}\cdot H_{L}\cdot\frac{\partial L}{\partial W}+\left(m\odot W\right)^{T}\cdot H_{L}\cdot\frac{\partial L}{\partial W}
\end{equation}is approximated via the first order Taylor series, with the Hessian
$H_{L}:=\frac{\partial^{2}L}{\partial W^{2}}\in\mathbb{R}^{D\times D}$.
As $\Delta L(W)-W^{T}\cdot H_{L}\cdot\frac{\partial L}{\partial W}$
is fix, the \emph{importance score} for all weights $W\in\mathbb{R}^{D}$
is computed as\begin{equation}\label{eq:importance_score}
S(W):=W\odot\left(H_{L}\cdot\frac{\partial L}{\partial W}\right)
\end{equation}and the weights with the $1-q$ highest importance scores are trained,
the other ones are pruned. Contrarily to the saliency score (\ref{eq:grad_saliency}),
the importance score \eqref{eq:importance_score} takes the sign into
account. Pruning with the GraSP criterion is applied, as SNIP and
FreezeNet, once before the training starts \cite{wang_2020}. Thus,
GraSP pruned networks also have sparse gradients during training. 

\subsubsection{Training Setup for Result in Table \ref{tab:Comparison-caffe}}
Different freezing rates are displayed in Table \ref{tab:Comparison-caffe}, as we first tried to find results in the corresponding papers with freezing rates approximately $q=0.99$. If no result for such a freezing rate was given, we report the result with the closest accuracy to FreezeNet's performance for $q=0.99$. 

We used the experimental setup described in Section \ref{sec:Experiments}
with hyperparameters from Section \ref{subsec:Experimental-Setup}.
The official method to calculate the GraSP score was used.\footnote{https://github.com/alecwangcq/GraSP.}
Moreover, learning rates $2^{n}\cdot0.1$ with $n\in\{-4,\ldots,4\}$
were tested for a split of training/validation images of $9/1$, $19/1$
and $3/1$. Each of these $27$ combinations was run $5$ times with
a different initialization. For each combination, the network with
the best validation accuracy was tested at its early stopping time.
The best result was reported for the learning rate $\lambda=0.1$
with a split of training/validation images of $19$/$1$ — the same
setup as used for the best FreezeNet result. The reported test accuracy
for the GraSP pruned network is given by $98.9\%$.

\subsection{Reinitialization Tests}

Up to now it is unanswered how the reinitialization of a network's
weights after the calculation of the saliency scores affects the trainability
of this network. To check this, we modify Algorithm \ref{alg:RSNIP}
by adding a reinitialization step after the computation of the saliency
score. FreezeNets with reinitializations are introduced in Algorithm
\ref{alg:GreeNet-with-Reinitialization}. 
\begin{algorithm}[tb]
	\begin{algorithmic}[1]
		\Require{Freezing rate $q$, initial parametrization $\Theta_0=W_0\cup B_0$, reinitialized parameters $\Theta_1 = W_1 \cup B_1$, network $f$, loss function $L$} 
		\Statex
		\State{Calculate saliency score $g \in \mathbb{R}^{\vert W_0 \vert}$ according to equation (\ref{eq:grad_saliency}) with $f_{\Theta_0}$ for one training batch}
		\State{Define freezing mask $m \in \mathbb{R}^{\vert W_0 \vert}$}
		\State{Calculate freezing threshold $\varepsilon$, where $\varepsilon$ denotes the $\lfloor (1 - q) \cdot \vert W_0 \vert \rfloor$-highest magnitude of $g$}
		\State{Set $m_k = 0$ if $\vert g_k \vert < \varepsilon$ else $m_k = 1$}
		\State{Set networks parameters to $\Theta_1$, i.e. use $f_{\Theta_1}$ for training}
		\State{Start training with forward propagation as usual but backpropagate gradient ${m \odot \frac{\partial L}{\partial W_1}}$} for weights and $\frac{\partial L}{\partial B_1}$ for biases
	\end{algorithmic}
	
	\caption{FreezeNet with Reinitialization after Computation of Freezing Mask\label{alg:GreeNet-with-Reinitialization}}
\end{algorithm}
We have tested a LeNet-$5$-Caffe baseline architecture for MNIST on Algorithm
\ref{alg:GreeNet-with-Reinitialization}. Again we follow the training
setup from Sections \ref{sec:Experiments} together with \ref{subsec:Experimental-Setup}.

\begin{figure}[tb]
	\begin{centering}
		\includegraphics[width=0.5\textwidth]{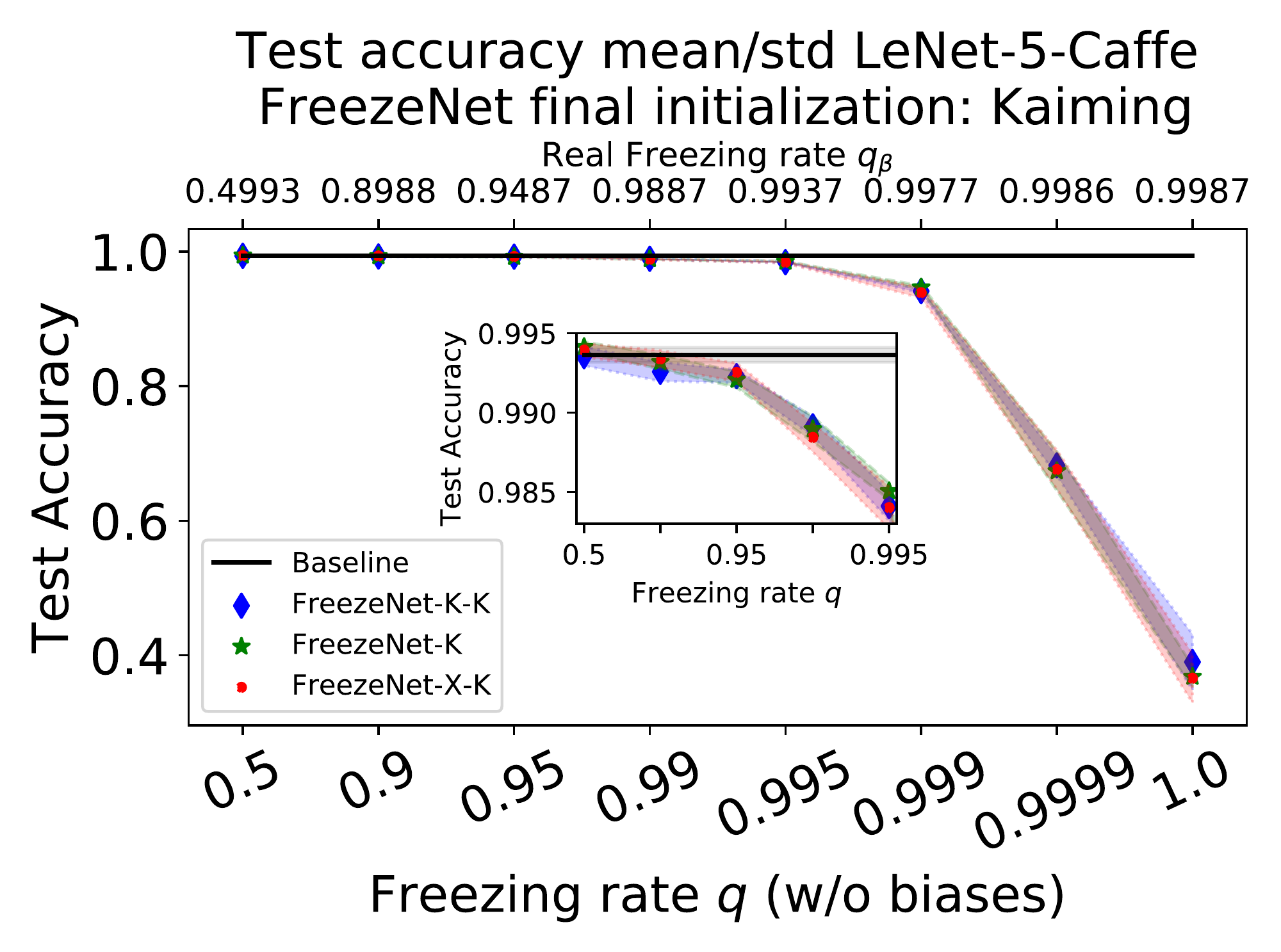}\includegraphics[width=0.5\textwidth]{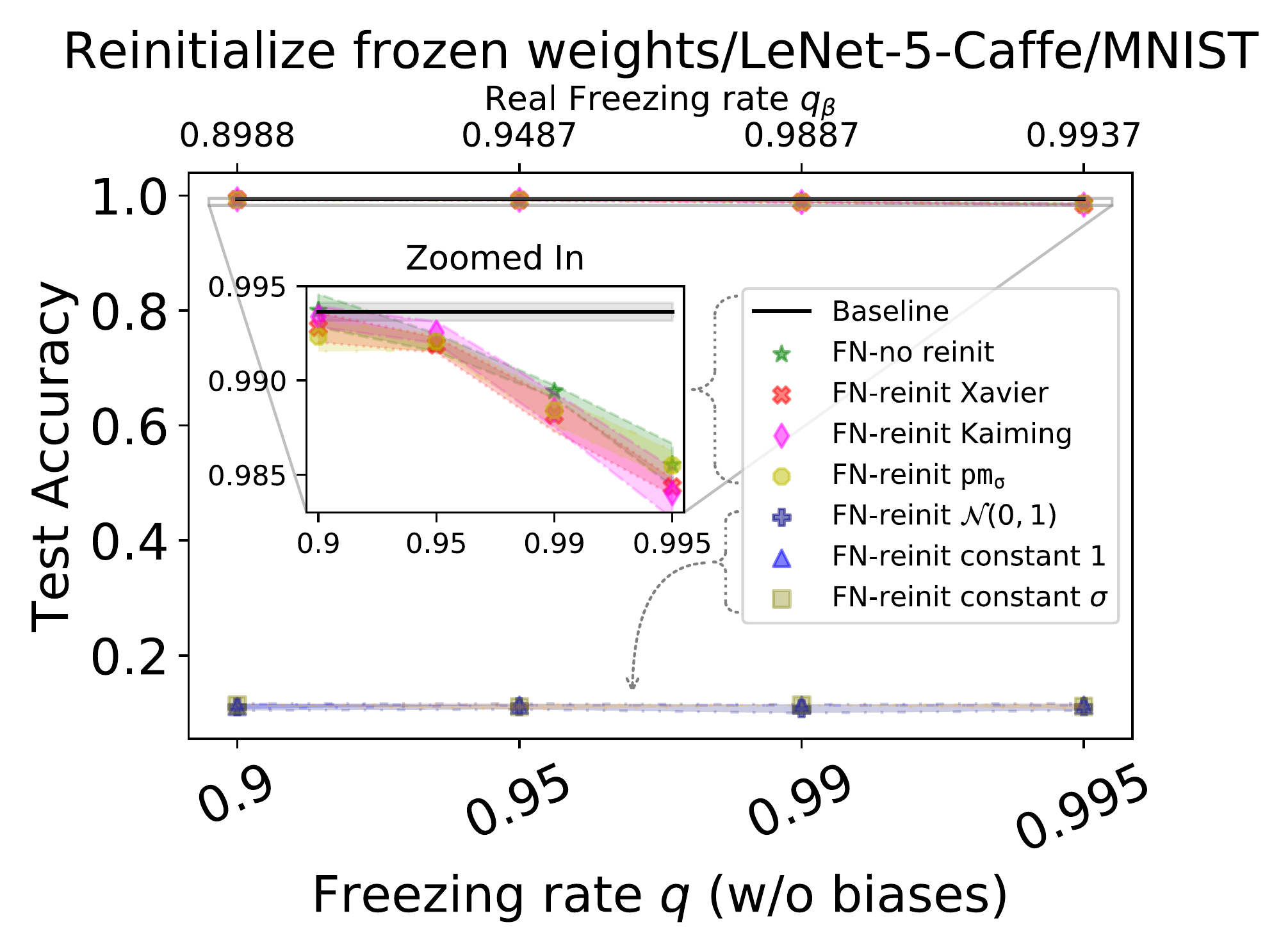}
		\par\end{centering}
	\caption{\label{fig:reinit_test2}Left: LeNet-$5$-Caffe baseline architecture with
		a corresponding FreezeNet-K, a FreezeNet-K-K and a FreezeNet-X-K. Right: Comparison of different reinitializations for a FreezeNet (FN) with LeNet-$5$-Caffe baseline architecture with Xavier-normal initialization.
		Inserted: Zoomed in versions of the plots.}
\end{figure}

First we tested combinations of initializing and reinitializing a FreezeNet with the Xavier-normal and the Kaiming-uniform initialization schemes.
We name the network without reinitialized weights FreezeNet-K for
a Kaiming-uniform initialized network, or FreezeNet-X for a Xavier-normal initialized one. The reinitialized networks are denoted as
FreezeNet-A-B for $\text{A,B}\in\{\text{K,X}\}$ where A stands for
the initialization used for finding the freezing mask and B for the
reinitialization, applied before training. 
\begin{figure}[tb]
	\includegraphics[width=0.5\textwidth]{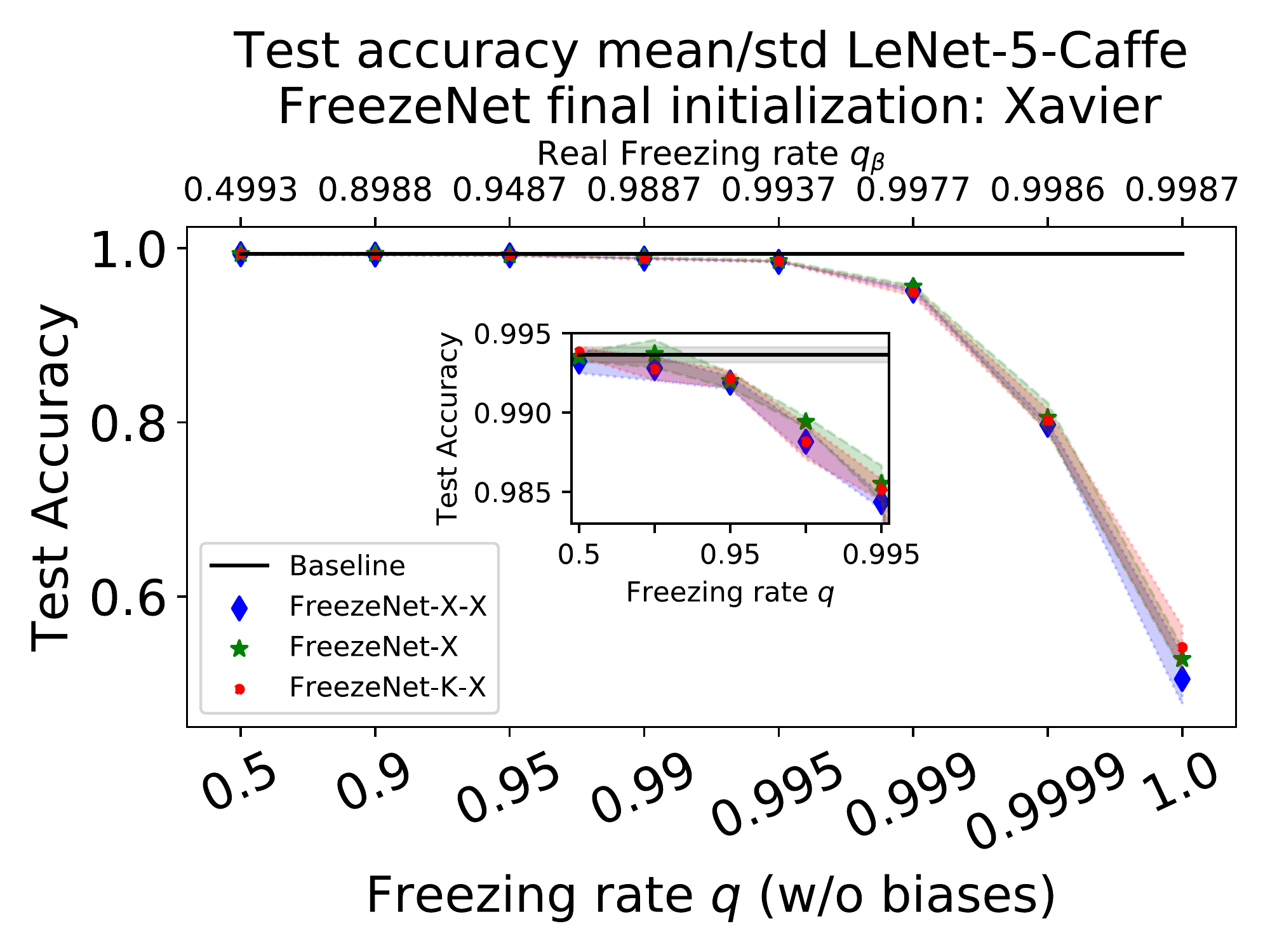}\includegraphics[width=0.5\textwidth]{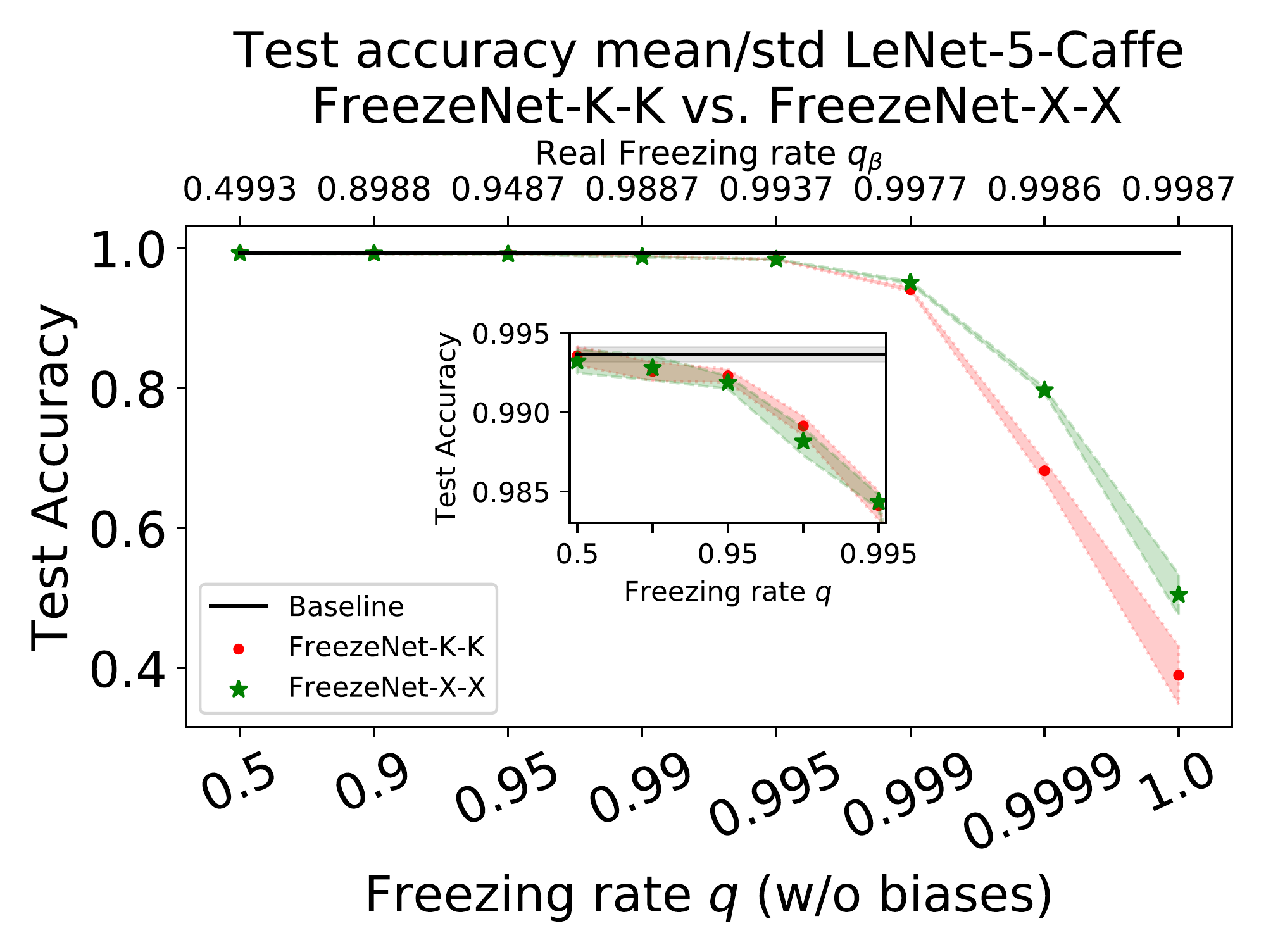}
	
	\caption{\label{fig:reinit_test}LeNet-$5$-Caffe baseline architecture compared
		to a FreezeNet-X, a FreezeNet-X-X and a FreezeNet-K-X in the left
		plot. On the right side we compare a FreezeNet-X-X with a FreezeNet-K-K.
		Inserted: Zoomed in versions of the plots.}
\end{figure}
The left plot in Figure \ref{fig:reinit_test} compares FreezeNet-X
with FreezeNet-X-X and FreezeNet-K-X. This graph shows, that reinitializations
do not significantly change the performance of FreezeNets. It also does not seem to make a difference if the probability distribution
used for the reinitialization differs from the one used to calculate
the freezing mask, examined through FreezeNet-K-X and FreezeNet-X-X.
Similar results are reported for the comparison of FreezeNet-K with
FreezeNet-K-K and FreezeNet-X-K, shown in the left part of Figure \ref{fig:reinit_test2}.

In Figure \ref{fig:reinit_test2}, right plot, various other reinitialization methods are tested on a Xavier-normal initialized FreezeNet.
We can see, that the variance scaling reinitialization methods Xavier-normal, Kaiming-uniform and ${\tt pm}_\sigma$ lead to the same results. The ${\tt pm}_\sigma$ initialization scheme is introduced in Section \ref{sec:Discussions} in the main body of the text.
Using other, non-variance scaling methods as constant reinitialization (either with all values $1$ or the layers variance $\sigma$) or drawing weights i.i.d. from $\mathcal{N}(0,1)$ generates networks which cannot be trained at all.

The right plot in Figure \ref{fig:reinit_test} shows that the initialization,
used after the freezing mask is computed, is important to solve the
MNIST task successfully for high freezing rates. The Kaiming init- and reinitialization, FreezeNet-K-K,
performs slightly better for the lower freezing rates and is outperformed
by FreezeNet-X-X for the higher ones. Without reinitialization, a
similar behaviour for low and high freezing rates can be seen in Figure
\ref{fig:capcity_init} — right plot. 

Summarized, using an appropriate initialization scheme for the weights,
after the freezing mask is computed, is essential for achieving good
results with FreezeNets. Based on our experiments we suggest initializing (and reinitializing, if wanted) FreezeNets with variance scaling methods.

\FloatBarrier

\subsection{Learning Rate Experiments}

\begin{figure}[tb]
	\includegraphics[width=0.5\textwidth]{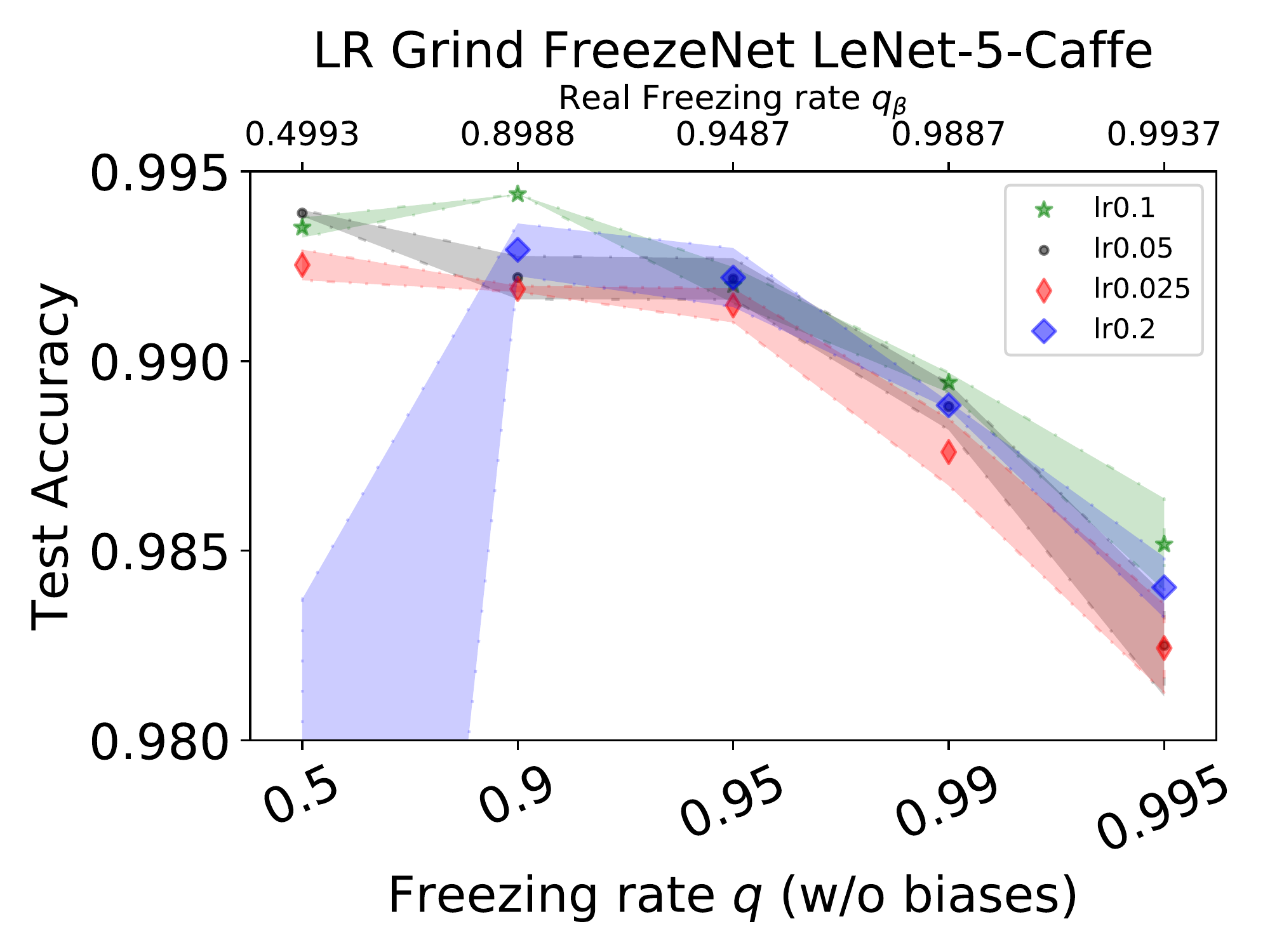}\includegraphics[width=0.5\textwidth]{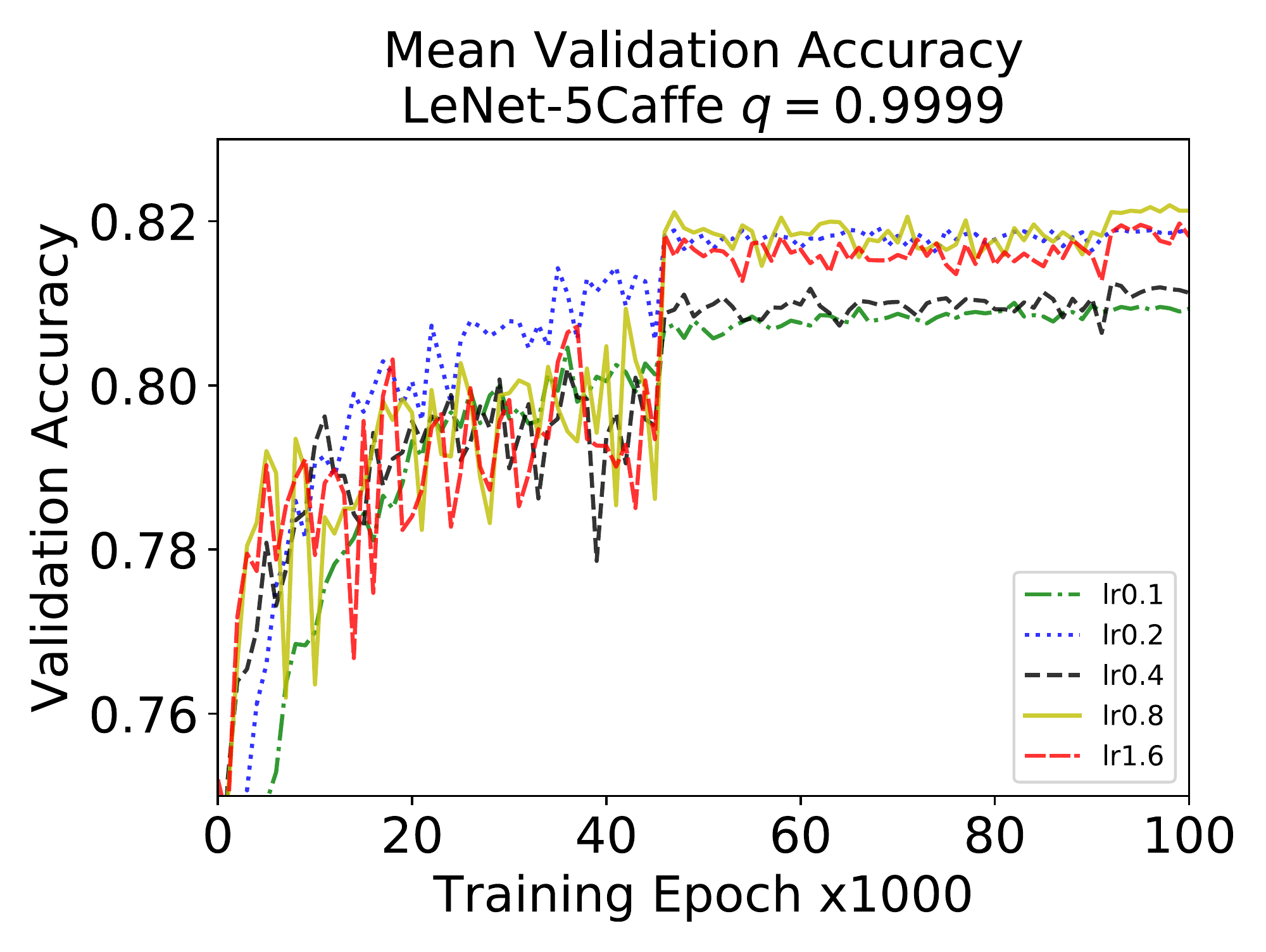}
	
	\caption{\label{fig:Learning-rate-tests}Learning rate tests for a FreezeNet
		with a LeNet-$5$-Caffe baseline. The left part shows the mean test
		accuracy for the best performing learning rates. The right plot shows
		the mean validation accuracy for three training runs recorded over
		the first $100$k epochs for different learning rates.}
\end{figure}

\begin{figure}[tb]
	\includegraphics[width=0.5\textwidth]{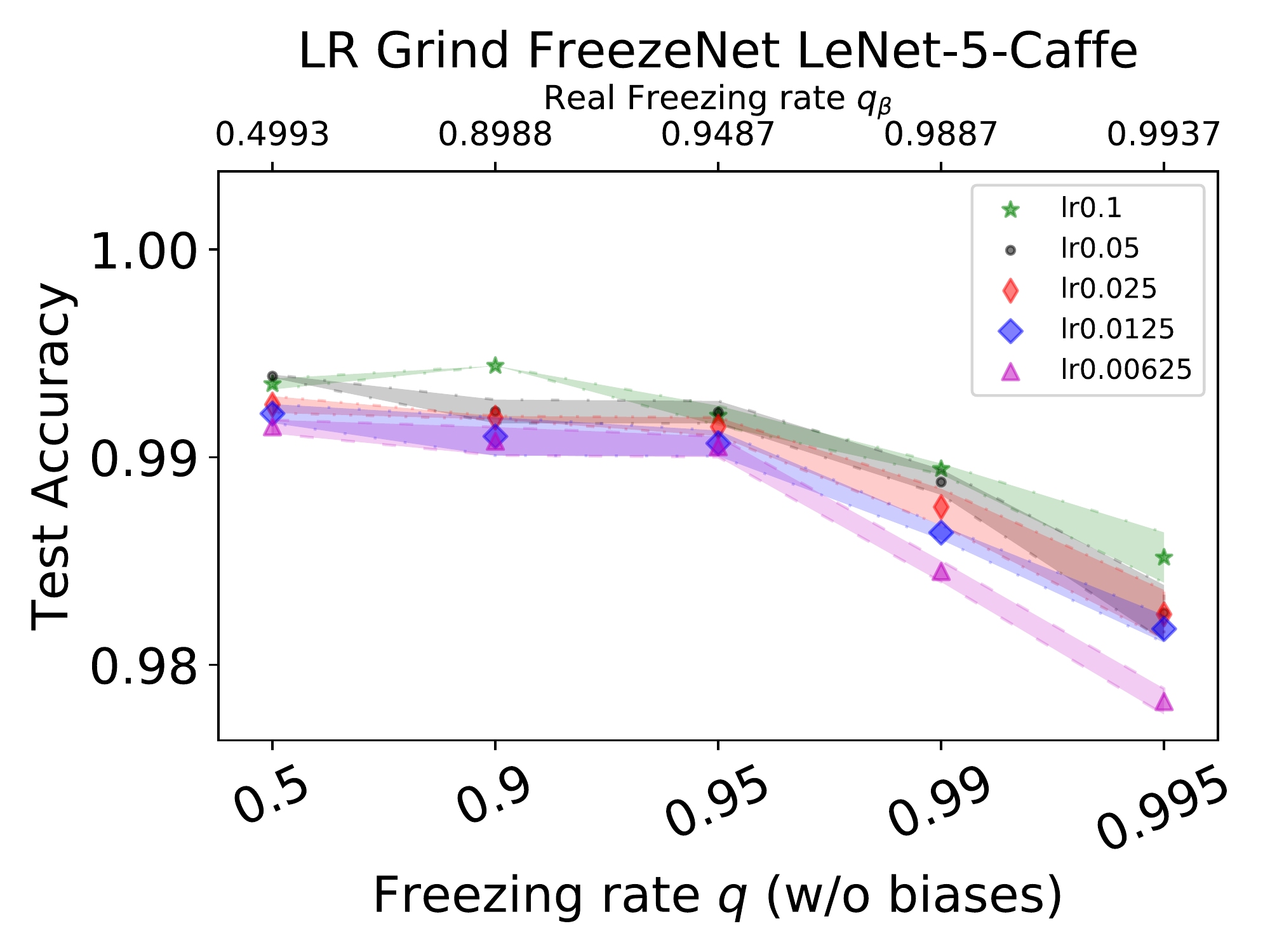}\includegraphics[width=0.5\textwidth]{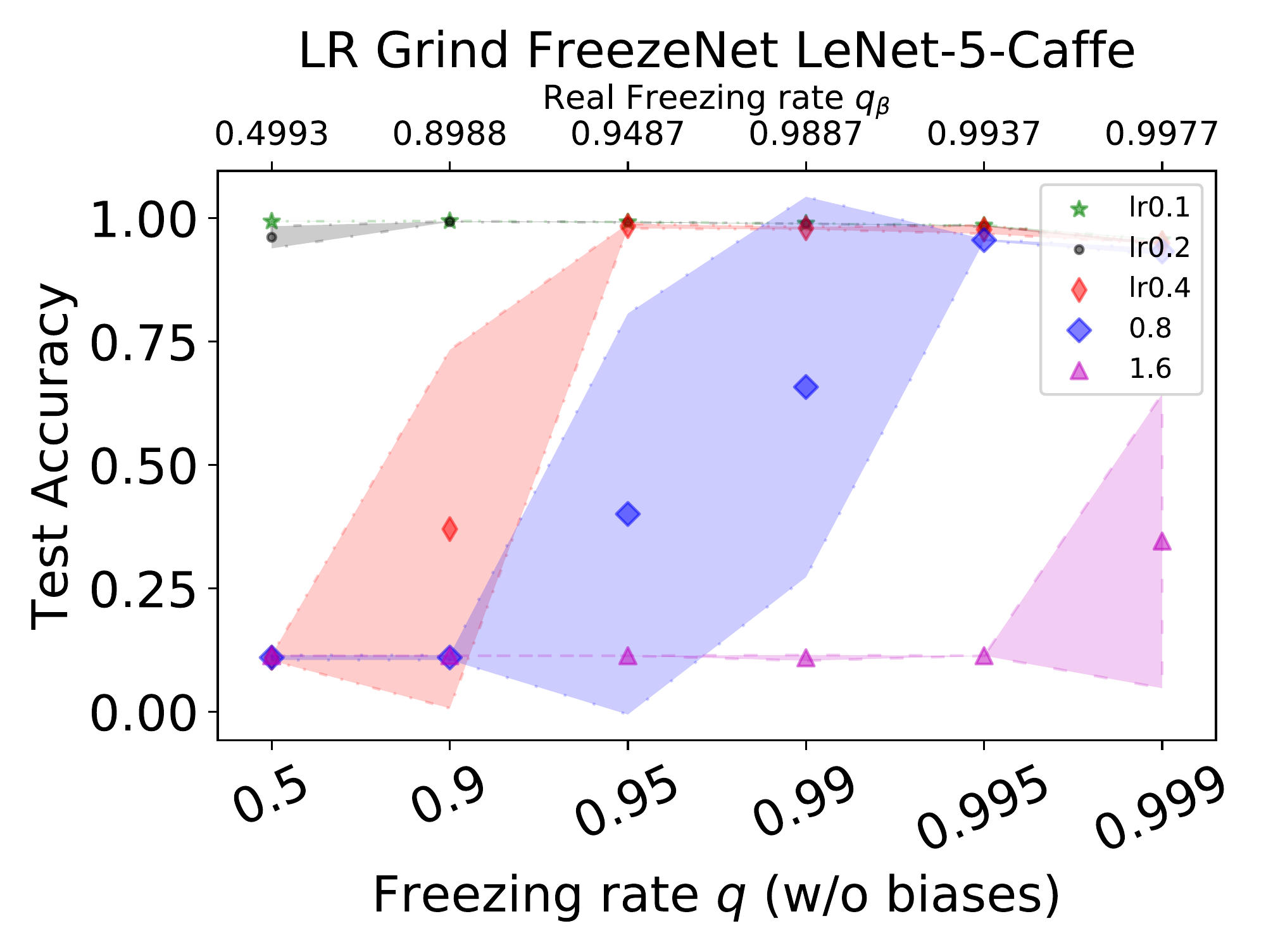}
	
	\caption{\label{fig:all_lr_test}Learning rate tests for a FreezeNet with a
		LeNet-$5$-Caffe baseline. The left part shows the mean test accuracy
		for the learning rates $\lambda\protect\leq\text{0.1}$. The right
		plot shows the mean validation accuracy for the learning rates $\lambda\protect\geq0.1$.}
\end{figure}

SGD is used in our experiments, thus we also tested a broad range
of learning rates for different freezing rates. This test was done
with the same setup as described in Sections \ref{sec:Experiments}
and \ref{subsec:Experimental-Setup}. The results for a LeNet-$5$-Caffe
baseline architecture on the MNIST classification task are shown in
Figures \ref{fig:Learning-rate-tests} and \ref{fig:all_lr_test}.
In order to cover a broad range of learning rates, we used $2^{n}\cdot\lambda_{0},n\in\{-4,-3,\ldots,4\}$
and $\lambda_{0}=0.1$ as learning rates. All learning and freezing
rate combinations were trained with three different random initializations.
The learning rates with the best results are shown in the left part
of Figure \ref{fig:Learning-rate-tests}. For almost any freezing
rate, the learning rate $0.1$ works best. Thus, FreezeNets do not
need expensive learning rate searches but can use a learning rate
performing well on the baseline architecture. But optimizing the learning
rate can improve FreeNets' performance beyond question. Another conclusion
we want to highlight is that higher learning
rates can be applied for higher freezing rates. Even if they do not work well for smaller freezing
rates, as the example of $\lambda=0.2$ in the left part of Figure
\ref{fig:Learning-rate-tests} shows. The same holds for learning
rates bigger than $0.2$, which require even higher freezing rates
to lead to satisfying results, as shown in the right part of Figure
\ref{fig:all_lr_test}. For high freezing rates, using higher learning
rates can lead to better and faster training, as shown in Figure \ref{fig:Learning-rate-tests},
right side. 

\begin{figure}[tb]
	\begin{centering}
		\includegraphics[width=0.5\textwidth]{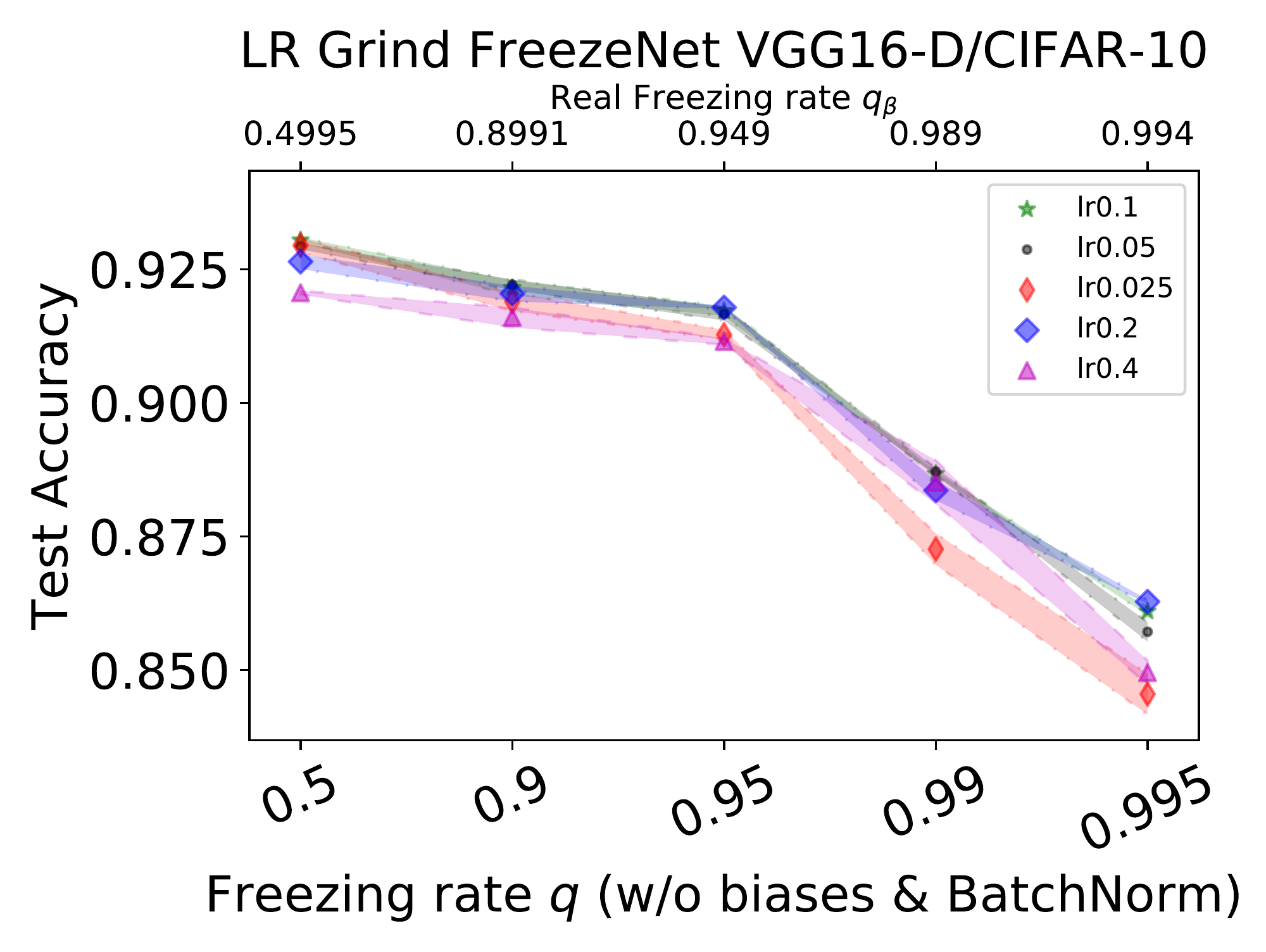}
		\par\end{centering}
	\caption{\label{fig:Lr_cifar10}Best performing learning rates for a FreezeNet
		with a VGG$16$-D baseline on the CIFAR-$10$ classification task. }
	
\end{figure}

The results for the learning rate search for the CIFAR\-$10$ task
with a VGG$16$-D are shown in Figure \ref{fig:Lr_cifar10}. Again,
$\lambda_{0}=0.1$ performs best for most of the freezing rates and
is only slightly improved for some freezing rates by $\lambda=0.2$
and $\lambda=0.05$.


\FloatBarrier

\subsection{\label{subsec:Figures-for-LeNet--Caffe}Figures for LeNet-$5$-Caffe on MNIST}

\begin{figure}[tb]
	\begin{centering}
		\includegraphics[width=0.5\textwidth]{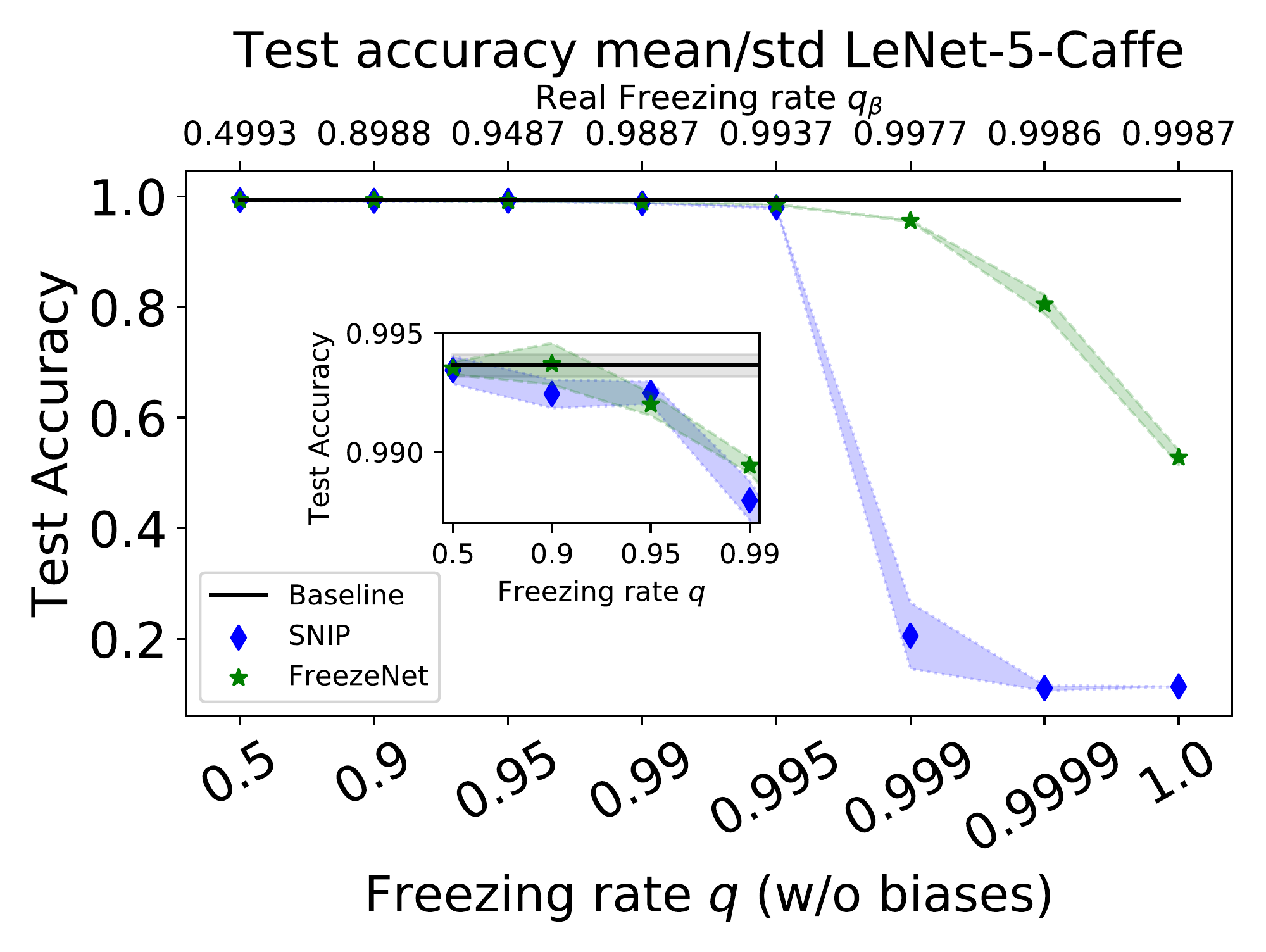}
		\par\end{centering}
	\caption{\label{fig:mnist_suppl}Comparison SNIP and FreezeNet for the MNIST
		classification task and a LeNet-$5$-Caffe baseline architecture.
		The inserted plot is a zoomed version.}
\end{figure}

Figure \ref{fig:mnist_suppl} shows the comparison of FreezeNet and
SNIP over a broad range of freezing rates, discussed in Section \ref{subsec:MNIST-with-LeNet--Caffe}. Here, the networks are trained and evaluated as described in Section
\ref{sec:Experiments} with hyperparameters from Section \ref{subsec:Experimental-Setup}.

\begin{figure}[tb]
	\begin{centering}
		\includegraphics[width=0.5\textwidth]{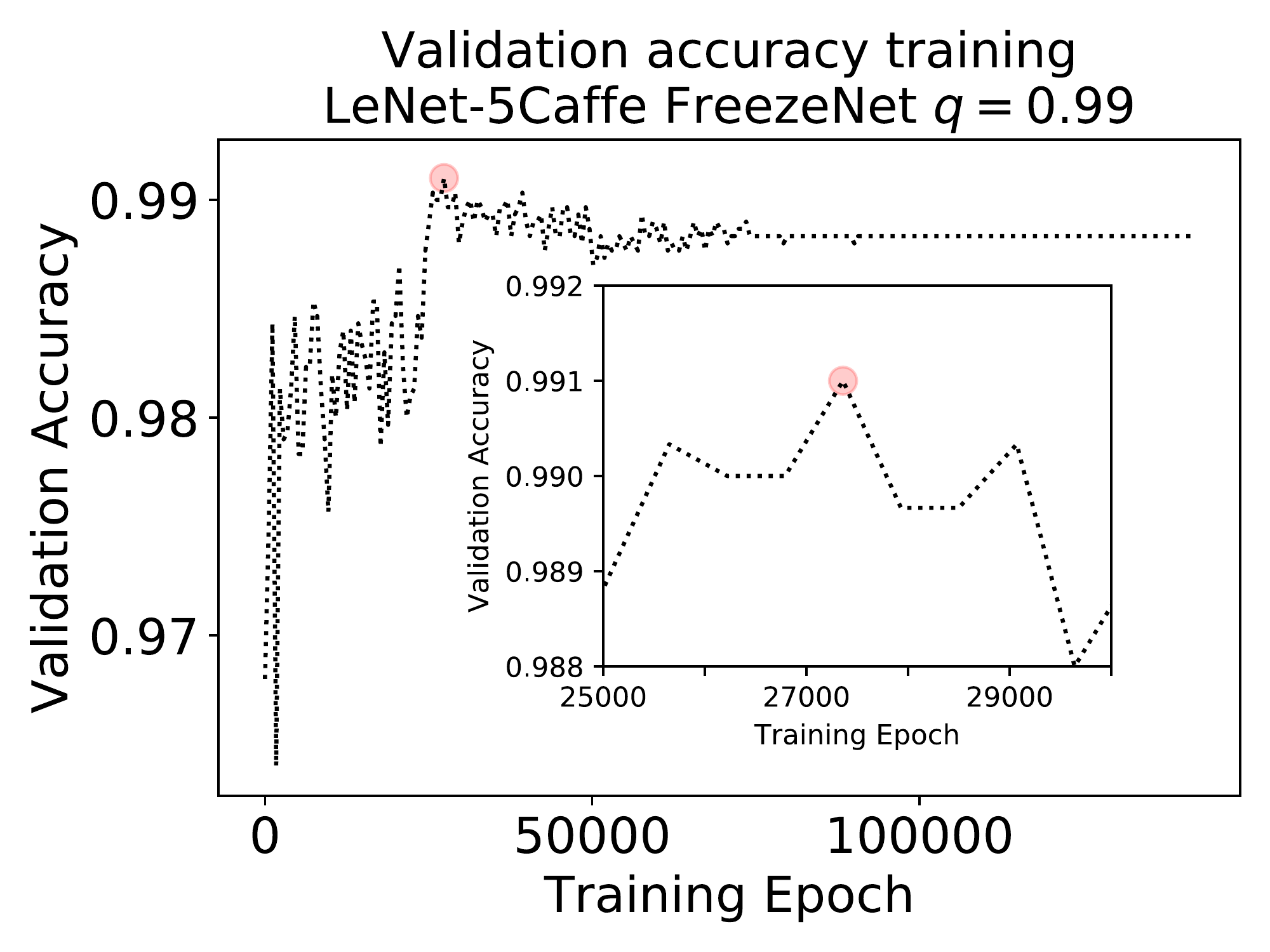}
		\par\end{centering}
	\caption{\label{fig:best_run}Training of FreezeNet for freezing rate $q=0.99$
		and baseline architecture LeNet-$5$-Caffe. Training procedure is
		done as described in Section \ref{subsec:Experimental-Setup} with hyperparameters from Section \ref{subsec:Experimental-Setup} but
		with a split $19/1$ of training and validation images. This plot
		shows the run with the best validation accuracy out of five tries.
		The corresponding test accuracy equals $99.1\%$, calculated with
		the weights stored in the early stop epoch, as reported in Table \ref{tab:Comparison-caffe}.
		The red circle highlights the epoch where early stopping occurs. The
		inserted plot is a zoomed version showing the early stopping epoch.}
\end{figure}

The training progress of FreezeNet's result, reported in Table \ref{tab:Comparison-caffe},
is shown in Figure \ref{fig:best_run}.

\subsection{Network Architectures}

Figures \ref{fig:Architecture_lenet300100}, \ref{fig:Architecture-caffe}
and \ref{fig:Architecture-vgg.} visualize the used LeNet-$300$-$100$,
LeNet-$5$-Caffe and VGG$16$-D network architectures, respectively. The ResNet$34$ architecture can be looked up in Table \ref{tab:ResNet} together with Figure \ref{fig:res_block}. 
\begin{figure}[tb]
	\tikzset{My Style/.style={draw, black, rectangle, minimum size=.5cm, minimum width=4.5cm, node distance = 0pt}}
	\tikzset{My Style2/.style={rectangle, minimum size=.5cm, minimum width=5.5cm, node distance = 0pt}}
	\tikzset{My Header/.style={minimum size=.5cm, minimum width=5.5cm, node distance = 0pt}}
	
	\begin{center}
		\begin{adjustbox}{width=.65\textwidth}
			\begin{tikzpicture}
			\node[My Style, fill=blue!20] (img) {Image $x \in \mathbb{R}^{1 \times 28 \times 28}$};
			\node[My Style2, below=of img] (arrow1) {};
			\node[My Style2, below=of arrow1] (arrow2) {};
			\node[My Style, below=of arrow2, fill=green!10] (input1) {Linear, $n_{in}=784$, $n_{out}=300$}; 
			\draw[->] (img.south) -- (input1.north);
			\node[My Style, below=of input1, fill=gray!20] (input2) {ReLU}; 
			
			\node[My Style, below=of input2, fill=green!10] (input3) {Linear, $n_{in}=300$, $n_{out}=100$};
			\node[My Style, below=of input3, fill=gray!20] (input4) {ReLU}; 
			
			\node[My Style, below=of input4, fill=green!10] (input5) {Linear $n_{in} = 100$, $n_{out}=10$};
			
			\node[My Style2, below=of input5] (input6) {};
			\node[My Style, below=of input6, fill=brown!10] (input7) {log softmax};
			\node[My Style2, below=of input7] (input8) {};
			\node[My Style, below=of input8, fill=blue!10] (input9) {Probability Vector $p \in [0,1]^{10}$};
			\draw[-] (input5.south) -- (input7.north) ;
			\draw[->] (input7.south) -- (input9.north) ;
			
			\draw[-, decorate,decoration={brace, amplitude=12pt}] (input1.east)  -- (input5.east)  node[midway, align=center, xshift=35pt, rotate=270]{Total Parameter count: \\ $266,610$ \\ ($410$ Bias)};
			
			
			
			

			\node[My Header,above=25pt of img] (t1) {\large \textbf {LeNet-}$\bm{300}$\textbf{-}$\bm{100}$\textbf{ Architecture}};
			
			\end{tikzpicture}
		\end{adjustbox}
	\end{center}
	
	\caption{\label{fig:Architecture_lenet300100}Architecture of the used LeNet-$300$-$100$.
		In front of the first layer, the feature map $x\in\mathbb{R}^{1\times28\times28}$
		is flattened to $\hat{x}\in\mathbb{R}^{784}$. For linear layers,
		$n_{in}$ and $n_{out}$ denote the number of incoming and outgoing
		neurons, respectively.}
\end{figure}
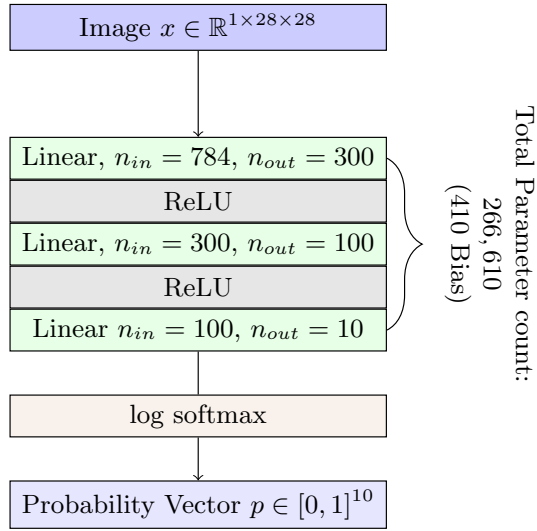
\begin{figure}[tb]
	\tikzset{My Style/.style={draw, black, rectangle, minimum size=.5cm, minimum width=4.5cm, node distance = 0pt}}
	\tikzset{My Style2/.style={rectangle, minimum size=.5cm, minimum width=5.5cm, node distance = 0pt}}
	\tikzset{My Header/.style={minimum size=.5cm, minimum width=5.5cm, node distance = 0pt}}
	\begin{center}
		\begin{adjustbox}{height=.75\textwidth}
			\begin{tikzpicture}
			\node[My Style, fill=blue!20] (img) {Image $x \in \mathbb{R}^{1 \times 28 \times 28}$};
			\node[My Style2, below=of img] (arrow1) {};
			\node[My Style2, below=of arrow1] (arrow2) {};
			\node[My Style, below=of arrow2, fill=gray!10] (input1) {Conv2D, $n_{in}=1$, $n_{out}=20$}; 
			\draw[->] (img.south) -- (input1.north);
			\node[My Style, below=of input1, fill=gray!20] (input2) {ReLU}; 
			\node[My Style, below=of input2, fill=brown!20] (input3) {MaxPool2D}; 
			\node[My Style, below=of input3, fill=gray!10] (input4) {Conv2D, $n_{in}=20$, $n_{out}=50$};
			\node[My Style, below=of input4, fill=gray!20] (input5) {ReLU}; 
			\node[My Style, below=of input5, fill=brown!20] (input6) {MaxPool2D}; 
			\node[My Style, below=of input6, fill=green!10] (input7) {Linear $n_{in} = 800$, $n_{out}=500$};
			\node[My Style, below=of input7, fill=gray!20] (input8) {ReLU};
			\node[My Style, below=of input8, fill=green!10] (input9) {Linear $n_{in} = 500$, $n_{out}=10$};
			\node[My Style2, below=of input9] (input10) {};
			\node[My Style, below=of input10, fill=brown!10] (input11) {log softmax};
			\node[My Style2, below=of input11] (input12) {};
			\node[My Style, below=of input12, fill=blue!10] (input13) {Probability Vector $p \in [0,1]^{10}$};
			\draw[-] (input9.south) -- (input11.north);
			\draw[->] (input11.south) -- (input13.north); 
			
			\draw[-, decorate,decoration={brace, amplitude=12pt}] (input1.east)  -- (input9.east)  node[midway, align=center, xshift=35pt, rotate=270]{Total Parameter count: \\ $431,080$ \\ ($580$ Bias)};
			
			\draw[->] (input1.south west) -- ++(-6pt, 0)  node[midway, left, yshift=0pt, xshift=-4pt, rotate=0]{ $24 \times 24$};
			
			\draw[->] (input3.south west) -- ++(-6pt, 0)  node[midway, left, yshift=0pt, xshift=-4pt, rotate=0]{ $12 \times 12$};
			
			\draw[->] (input4.south west) -- ++(-6pt, 0)  node[midway, left, yshift=0pt, xshift=-4pt, rotate=0]{ $8 \times 8$};
			
			\draw[->] (input6.south west) -- ++(-6pt, 0)  node[midway, left, yshift=0pt, xshift=-4pt, rotate=0]{ $4 \times 4$};
			
			\node[My Header,above=25pt of img] (t1) {\large \textbf {LeNet-}$\bm{5}$\textbf{-Caffe Architecture}};

			\end{tikzpicture}
		\end{adjustbox}
	\end{center}
	
	\caption{\label{fig:Architecture-caffe}Architecture of the used LeNet-$5$-Caffe.
		All 2D-convolutional layers have kernel size $5\times5$, $1\times1$
		stride and no zero padding. A max-pooling layer has kernel size $2\times2$,
		stride $2\times2$ and dilation $1\times1$. Before entering the first
		linear layer, the feature map $\hat{x}\in\mathbb{R}^{50\times4\times4}$
		is flattened to $\hat{\hat{x}}\in\mathbb{R}^{800}$. The resolutions
		left to the blocks denote the resolution of the feature maps, processed
		by the corresponding layers. For convolutional layers, $n_{in}$ and
		$n_{out}$ denote the number of incoming and outgoing channels, respectively.
		For linear layers, $n_{in}$ and $n_{out}$ denote the number of incoming
		and outgoing neurons, respectively.}
\end{figure}
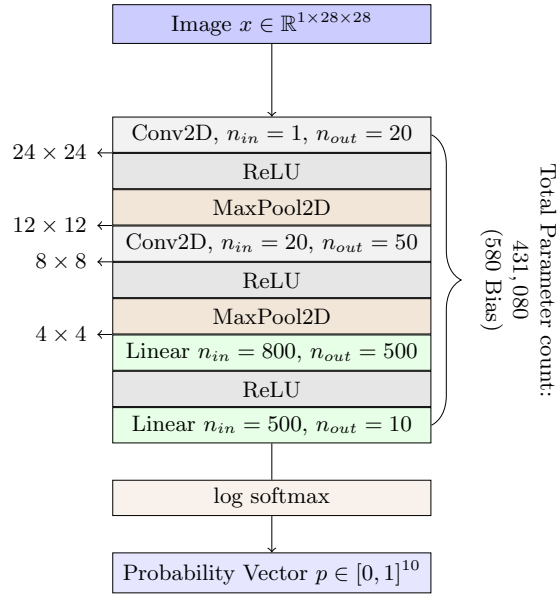
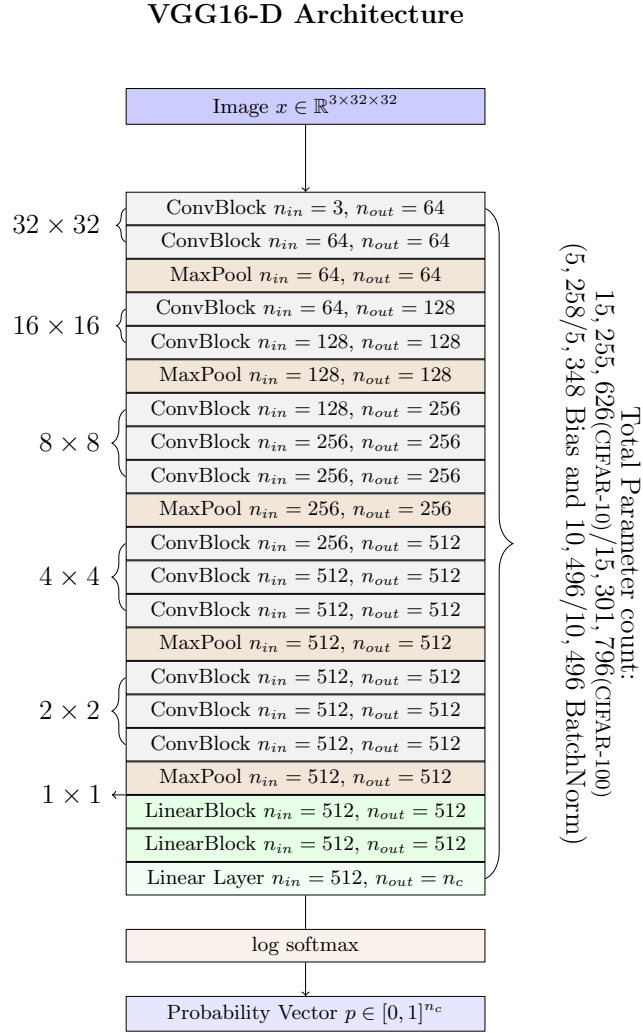
\begin{figure}[tb]
	\tikzset{My Style/.style={draw, black, rectangle, minimum size=.5cm, minimum width=5.5cm, node distance = 0pt}}
	\tikzset{My Style2/.style={rectangle, minimum size=.5cm, minimum width=5.5cm, node distance = 0pt}}
	\tikzset{My Header/.style={minimum size=.5cm, minimum width=5.5cm, node distance = 0pt}}
	
	\begin{center}
		\begin{adjustbox}{width=.7\textwidth}
			\begin{tikzpicture}
			\node[My Style, fill=blue!20] (img) {Image $x \in \mathbb{R}^{3 \times 32 \times 32}$};
			\node[My Style2, below=of img] (arrow1) {};
			\node[My Style2, below=of arrow1] (arrow2) {};
			\node[My Style, below=of arrow2, fill=gray!10] (input) {ConvBlock $n_{in}=3$, $n_{out}=64$}; 
			\draw[->] (img.south) -- (input.north);
			\node[My Style, below=of input, fill=gray!10] (input2) {ConvBlock $n_{in}=64$, $n_{out}=64$};
			\node[My Style, below=of input2, fill=brown!20] (input3) {MaxPool $n_{in}=64$, $n_{out}=64$};
			\node[My Style, below=of input3, fill=gray!10] (input4) {ConvBlock $n_{in}=64$, $n_{out}=128$};
			\node[My Style, below=of input4, fill=gray!10] (input5) {ConvBlock $n_{in}=128$, $n_{out}=128$};
			\node[My Style, below=of input5, fill=brown!20] (input6) {MaxPool $n_{in}=128$, $n_{out}=128$};
			\node[My Style, below=of input6, fill=gray!10] (input7) {ConvBlock $n_{in}=128$, $n_{out}=256$};
			\node[My Style, below=of input7, fill=gray!10] (input8) {ConvBlock $n_{in}=256$, $n_{out}=256$};
			\node[My Style, below=of input8, fill=gray!10] (input9) {ConvBlock $n_{in}=256$, $n_{out}=256$};
			\node[My Style, below=of input9, fill=brown!20] (input10) {MaxPool $n_{in}=256$, $n_{out}=256$};
			\node[My Style, below=of input10, fill=gray!10] (input11) {ConvBlock $n_{in}=256$, $n_{out}=512$};
			\node[My Style, below=of input11, fill=gray!10] (input12) {ConvBlock $n_{in}=512$, $n_{out}=512$};
			\node[My Style, below=of input12, fill=gray!10] (input13) {ConvBlock $n_{in}=512$, $n_{out}=512$};
			\node[My Style, below=of input13, fill=brown!20] (input14) {MaxPool $n_{in}=512$, $n_{out}=512$};
			\node[My Style, below=of input14, fill=gray!10] (input15) {ConvBlock $n_{in}=512$, $n_{out}=512$};
			\node[My Style, below=of input15, fill=gray!10] (input16) {ConvBlock $n_{in}=512$, $n_{out}=512$};
			\node[My Style, below=of input16, fill=gray!10] (input17) {ConvBlock $n_{in}=512$, $n_{out}=512$};
			\node[My Style, below=of input17, fill=brown!20] (input18) {MaxPool $n_{in}=512$, $n_{out}=512$};
			\node[My Style, below=of input18, fill=green!10] (input19) {LinearBlock $n_{in} = 512$, $n_{out}=512$};
			\node[My Style, below=of input19, fill=green!10] (input20) {LinearBlock $n_{in} = 512$, $n_{out}=512$};
			\node[My Style, below=of input20, fill=green!5] (input21) {Linear Layer $n_{in} = 512$, $n_{out}=n_c$};
			\node[My Style2, below=of input21] (input22) {};
			\node[My Style, below=of input22, fill=brown!10] (input23) {log softmax};
			\node[My Style2, below=of input23] (input24) {};
			\node[My Style, below=of input24, fill=blue!10] (input25) {Probability Vector $p \in [0,1]^{n_c}$};
			\draw[-] (input21.south) -- (input23.north);
			\draw[->] (input23.south) -- (input25.north);
			
			\draw[-, decorate,decoration={brace, amplitude=12pt}] (input.east)  -- (input21.east)  node[midway, align=center, xshift=50pt, rotate=270]{\large Total Parameter count: \\ \large $15,255,626$\small (CIFAR-$10$)\large/$15,301,796$\small (CIFAR-$100$) \\ \large ($5,258$/$5,348$ Bias and $10,496$/$10,496$ BatchNorm)};
			
			\draw[-, decorate,decoration={brace, mirror, amplitude=4pt}] (input.west)  -- (input2.west)  node[midway, left,yshift=0pt, xshift=-8pt, rotate=0]{\large $32 \times 32$};
			
			\draw[-, decorate,decoration={brace, mirror, amplitude=4pt}] (input4.west)  -- (input5.west)  node[midway, left,yshift=0pt, xshift=-8pt, rotate=0]{\large $16 \times 16$};
			
			\draw[-, decorate,decoration={brace, mirror, amplitude=6pt}] (input7.west)  -- (input9.west)  node[midway, left,yshift=0pt, xshift=-8pt, rotate=0]{\large $8 \times 8$};
			
			\draw[-, decorate,decoration={brace, mirror, amplitude=6pt}] (input11.west)  -- (input13.west)  node[midway, left,yshift=0pt, xshift=-8pt, rotate=0]{\large $4 \times 4$};

			\draw[-, decorate,decoration={brace, mirror, amplitude=6pt}] (input15.west)  -- (input17.west)  node[midway, left,yshift=0pt, xshift=-8pt, rotate=0]{\large $2 \times 2$};
			
			\node[left=of input18] (dummy) {};
			\draw[->] (input18.south west) -- ++(-6pt, 0)  node[midway, left, yshift=0pt, xshift=-4pt, rotate=0]{\large $1 \times 1$};
			
			\node[My Header,above=25pt of img] (t1) {\large \textbf {VGG}$\bm{16}$\textbf{-D Architecture}};
			
			\end{tikzpicture}
		\end{adjustbox}
	\end{center}
	
	\caption{\label{fig:Architecture-vgg.}Architecture of VGG$16$-D. Here, $n_{c}\in\{10,100\}$
		equals the numbers of classes for the given classification task. For
		CIFAR-$10$, $n_{c}=10$ and for CIFAR-$100$ we have $n_{c}=100.$
		A ConvBlock consists of a 2D-convolutional layer with kernel size
		$3\times3$, and $1\times1$ stride and padding. Each convolution is
		followed by a $2$D-Batch Normalization Layer and a ReLU activation
		function. The max-pooling layer has kernel size $2\times2$, stride
		$2\times2$ and dilation $1\times1$. Before entering the first LinearBlock,
		the feature map $\hat{x}\in\mathbb{R}^{512\times1\times1}$ is flattened
		to $\hat{\hat{x}}\in\mathbb{R}^{512}$. A LinearBlock consist of a
		fully connected layer followed by a $1$D-Batch Normalization Layer
		and a ReLU activation function. The resolution left to the blocks
		denotes the resolution of the feature maps, processed by the corresponding
		blocks. For Convolutional blocks, $n_{in}$ and $n_{out}$ denote
		the number of incoming and outgoing channels, respectively. For linear
		blocks and layers, $n_{in}$ and $n_{out}$ denote the number of incoming and
		outgoing neurons, respectively.}
\end{figure}

\begin{table}
	\caption{\label{tab:ResNet}ResNet$34$ with \texttt{ResBlocks}, shown in
		Figure \ref{fig:res_block}. The kernel size is given by $k$. For Convolutional layers and \texttt{ResBlocks}, $n_{in}$ and $n_{out}$ denote
		the number of incoming and outgoing channels, respectively. For the linear
		layer, $n_{in}$ and $n_{out}$ denote the number of incoming and
		outgoing neurons, respectively. Before entering the linear layer,
		the feature map $x\in\mathbb{R}^{512\times1\times1}$ is flattened
		to $\hat{x}\in\mathbb{R}^{512}$. A ResNet$34$ consists of $21,383,816$ parameters in total. Thereof, $21,366,464$ weights, $200$ biases and $17,152$ BatchNorm parameters.}
	{\scriptsize{}}%
	\begin{tabular}{ccccccccccc}
		{\footnotesize{}Module} & {\footnotesize{}Output Size}  & {\footnotesize{}Repeat} & {\footnotesize{}$n_{in}$} & {\footnotesize{}$n_{out}$} & {\footnotesize{}$k$} & {\footnotesize{}Stride} & {\footnotesize{}Padding} & {\footnotesize{}Bias} & {\footnotesize{}BatchNorm} & {\footnotesize{}ReLU}\tabularnewline
		\hline 
		{\footnotesize{}Conv2D} & {\scriptsize{}$64\times64$} & {\scriptsize{}$\times 1$} &{\scriptsize{}$3$} & {\scriptsize{}$64$} & {\scriptsize{}$3$} & {\scriptsize{}$1$} & {\scriptsize{}$1$} & {\scriptsize{}\xmark} & {\scriptsize{}\cmark} & {\scriptsize{}\cmark}\tabularnewline
		{\footnotesize{}\texttt{ResBlock}} & {\scriptsize{}$32\times32$} &{\scriptsize{}$\times 1$} & {\scriptsize{}$64$} & {\scriptsize{}$64$} & {\scriptsize{}$3$} & {\scriptsize{}$2$} & {\scriptsize{}$1$} & {\scriptsize{}\xmark} & {\scriptsize{}\cmark} & {\scriptsize{}\cmark}\tabularnewline
		{\footnotesize{}\texttt{ResBlock}} & {\scriptsize{}$32\times32$} &{\scriptsize{}$\times 2$} & {\scriptsize{}$64$} & {\scriptsize{}$64$} & {\scriptsize{}$3$} & {\scriptsize{}$1$} & {\scriptsize{}$1$} & {\scriptsize{}\xmark} & {\scriptsize{}\cmark} & {\scriptsize{}\cmark}\tabularnewline		
		{\footnotesize{}\texttt{ResBlock}} & {\scriptsize{}$16\times16$} & {\scriptsize{}$\times 1$}&{\scriptsize{}$64$} & {\scriptsize{}$128$} & {\scriptsize{}$3$} & {\scriptsize{}$2$} & {\scriptsize{}$1$} & {\scriptsize{}\xmark} & {\scriptsize{}\cmark} & {\scriptsize{}\cmark}\tabularnewline
		{\footnotesize{}\texttt{ResBlock}} & {\scriptsize{}$16\times16$} & {\scriptsize{}$\times 3$}&{\scriptsize{}$128$} & {\scriptsize{}$128$} & {\scriptsize{}$3$} & {\scriptsize{}$1$} & {\scriptsize{}$1$} & {\scriptsize{}\xmark} & {\scriptsize{}\cmark} & {\scriptsize{}\cmark}\tabularnewline
		{\footnotesize{}\texttt{ResBlock}} & {\scriptsize{}$8\times8$} & {\scriptsize{}$\times 1$} &{\scriptsize{}$128$} & {\scriptsize{}$256$} & {\scriptsize{}$3$} & {\scriptsize{}$2$} & {\scriptsize{}$1$} & {\scriptsize{}\xmark} & {\scriptsize{}\cmark} & {\scriptsize{}\cmark}\tabularnewline
		{\footnotesize{}\texttt{ResBlock}} & {\scriptsize{}$8\times8$} & {\scriptsize{}$\times 5$} &{\scriptsize{}$256$} & {\scriptsize{}$256$} & {\scriptsize{}$3$} & {\scriptsize{}$1$} & {\scriptsize{}$1$} & {\scriptsize{}\xmark} & {\scriptsize{}\cmark} & {\scriptsize{}\cmark}\tabularnewline
		{\footnotesize{}\texttt{ResBlock}} & {\scriptsize{}$4\times4$} & {\scriptsize{}$\times 1$}&{\scriptsize{}$256$} & {\scriptsize{}$512$} & {\scriptsize{}$3$} & {\scriptsize{}$2$} & {\scriptsize{}$1$} & {\scriptsize{}\xmark} & {\scriptsize{}\cmark} & {\scriptsize{}\cmark}\tabularnewline
		{\footnotesize{}\texttt{ResBlock}} & {\scriptsize{}$4\times4$} & {\scriptsize{}$\times 2$}&{\scriptsize{}$512$} & {\scriptsize{}$512$} & {\scriptsize{}$3$} & {\scriptsize{}$1$} & {\scriptsize{}$1$} & {\scriptsize{}\xmark} & {\scriptsize{}\cmark} & {\scriptsize{}\cmark}\tabularnewline
		{\footnotesize{}AvgPool2D} & {\scriptsize{}$1\times1$} & {\scriptsize{}$\times 1$}&{\scriptsize{}$512$} & {\scriptsize{}$512$} & {\scriptsize{}$4$} & {\scriptsize{}$0$} & {\scriptsize{}$0$} & {\scriptsize{}\xmark} & {\scriptsize{}\xmark} & {\scriptsize{}\xmark}\tabularnewline
		{\footnotesize{}Linear} & {\scriptsize{}$200$} & {\scriptsize{}$\times 1$}&{\scriptsize{}$512$} & {\scriptsize{}$200$} & {\scriptsize{}---} & {\scriptsize{}---} & {\scriptsize{}---} & {\scriptsize{}\cmark} & {\scriptsize{}\xmark} & {\scriptsize{}\xmark}\tabularnewline
		{\footnotesize{}log softmax} & {\scriptsize{}$10$} & {\scriptsize{}$\times 1$}&{\scriptsize{}$200$} & {\scriptsize{}$200$} & {\scriptsize{}---} & {\scriptsize{}---} & {\scriptsize{}---} & {\scriptsize{}\xmark} & {\scriptsize{}\xmark} & {\scriptsize{}\xmark}\tabularnewline
	\end{tabular}
\end{table}
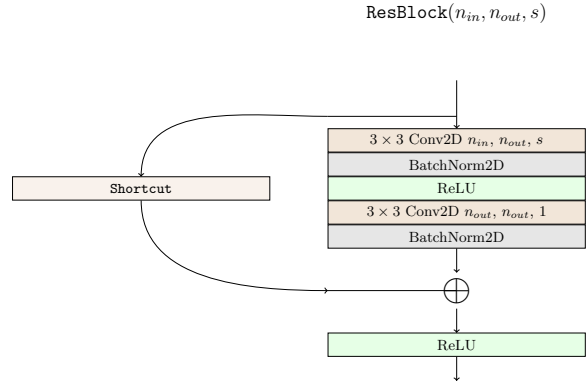
\begin{figure}
	\tikzset{My Style/.style={draw, black, rectangle, minimum size=.5cm, minimum width=5.5cm, node distance = 0pt}}
	\tikzset{My Style2/.style={rectangle, minimum size=.5cm, minimum width=5.5cm, node distance = 0pt}}
	\tikzset{My Header/.style={minimum size=.5cm, minimum width=6.5cm, node distance = 0pt}}
	
	\begin{center}
		\begin{adjustbox}{width=.65\textwidth}
			\begin{tikzpicture}
			\node (img) {};
			\node[My Style2, below=of img] (arrow1) {};
			\node[My Style2, below=of arrow1] (arrow2) {};
			\node[My Style, below=of arrow2, fill=brown!20] (input1) {$3\times 3$ Conv2D $n_{in}$, $n_{out}$, $s$}; 
			\draw[->] (img.south) -- (input1.north);
			
			\node[My Style, below=of input1, fill=gray!20] (input2) {BatchNorm2D}; 
			\node[My Style, below=of input2, fill=green!10] (input3) {ReLU};
			
			\node[My Style, below=of input3, fill=brown!20] (input4) {$3\times 3$ Conv2D $n_{out}$, $n_{out}$, $1$}; 
			\node[My Style, below=of input4, fill=gray!20] (input5) {BatchNorm2D};

			\node[My Style2, below=of input5] (arrow3) {};
			\node[My Style2, below=of arrow3] (arrow4) {\scalebox{2.5}{$\oplus$}};
			\node[My Style2, below=of arrow4] (arrow5) {};
			\node[My Style, below=of arrow5, fill=green!10] (input6) {ReLU};
			\node[My Style2, below=of input6] (arrow6) {};
			\node[left = of input3] (res_dum1) {};
			\node[My Style, left = of res_dum1, fill=brown!10] (res1) {\texttt{Shortcut}};
			
			\draw[->] (arrow2.west) to [out=180, in=90] (res1.north) {};
			\draw[-] (arrow2.west) -- (arrow2.center);
			
			\draw[->] (res1.south) to [out=-90, in=180] (arrow4.west) {};
			\draw[-] (arrow4.west) -- (arrow4.center);
			\draw[->] (input5.south) -- (arrow4.north) {}; 
			
			\draw[->] (arrow4.south) -- (input6.north) {}; 
			\draw[->] (input6.south) -- (arrow6.south) {};

			\node[My Header,above=25pt of img] (t1) {\large \textbf {\texttt{ResBlock}$(n_{in}, n_{out}, s)$}};
			\end{tikzpicture}
		\end{adjustbox}
	\end{center}
	
	\caption{\label{fig:res_block}Architecture of a \texttt{ResBlock} with $n_{in}$
		input channels, $n_{out}$ output channels and stride $s$
		for the first convolution. The second $3\times 3$ conolution has stride $1$ and $n_{out}$ in- and output channels.
		\texttt{Shortcut} connections are a $1\times1$
		2D Convolution with $n_{in}$ input and $n_{out}$ output channels and stride $s$ followed by a BatchNorm2D layer if $n_{in}\neq n_{out}$. Otherwise, \texttt{Shortcut} connections are simply given by the \texttt{Identity} function.
		All Conv2D layers are initialized without biases.}
\end{figure}
\end{document}